\documentclass[letterpaper]{article} % DO NOT CHANGE THIS
\usepackage{aaai2026}  % DO NOT CHANGE THIS
\usepackage{times}  % DO NOT CHANGE THIS
\usepackage{helvet}  % DO NOT CHANGE THIS
\usepackage{courier}  % DO NOT CHANGE THIS
\usepackage[hyphens]{url}  % DO NOT CHANGE THIS
\usepackage{graphicx} % DO NOT CHANGE THIS
\usepackage{natbib}  % DO NOT CHANGE THIS AND DO NOT ADD ANY OPTIONS TO IT
\usepackage{caption} % DO NOT CHANGE THIS AND DO NOT ADD ANY OPTIONS TO IT
\usepackage{algorithm}
\usepackage{algorithmic}

\usepackage{microtype}
\usepackage{graphicx}
\usepackage{subcaption}
\usepackage{booktabs}
\usepackage{amsmath}
\usepackage{xcolor}
\usepackage{tcolorbox}
\usepackage{amsmath}
\usepackage{amssymb}
\usepackage{mathtools}
\usepackage{amsthm}
\usepackage{enumitem}
\usepackage{multirow}
\usepackage{adjustbox}
\usepackage{longtable}
\usepackage{array}

\newtheorem{theorem}{Theorem}[section]

\newtheorem{lemma}[theorem]{Lemma}

\theoremstyle{definition}

\theoremstyle{remark}

\newtcolorbox{questionbox}{
    colback=blue!5,
    colframe=blue!50,
    boxrule=1pt,
    top=5pt,
    bottom=5pt,
    left=8pt,
    right=8pt,
    arc=4pt,
    boxsep=0pt,
    toptitle=1pt,
    bottomtitle=1pt,
    before skip=6pt,  % space above the box
    after skip=6pt,   % space below the box
}

\newcommand{\our}{\texttt{IPEA-HF}}

\usepackage{newfloat}
\usepackage{listings}
\DeclareCaptionStyle{ruled}{labelfont=normalfont,labelsep=colon,strut=off} % DO NOT CHANGE THIS
\floatstyle{ruled}
\newfloat{listing}{tb}{lst}{}
\floatname{listing}{Listing}
\title{Preference is More Than Comparisons: Rethinking Dueling Bandits with Augmented Human Feedback}
\author{
    Shengbo Wang\textsuperscript{\rm 1},
    Hong Sun\textsuperscript{\rm 1},
    Ke Li\textsuperscript{\rm 2}
}
\affiliations{
    \textsuperscript{\rm 1}School of Computer Science and Engineering, University of Electronic Science and Technology of China, Chengdu, China\\
    \textsuperscript{\rm 2}Department of Computer Science, University of Exeter, EX4 4RN, Exeter, UK\\
    shnbo.wang@foxmail.com; 
    k.li@exeter.ac.uk
}

\usepackage{bibentry}
\begin{document}

\maketitle

\begin{abstract}
Interactive preference elicitation (IPE) aims to substantially reduce human effort while acquiring human preferences in wide personalization systems. Dueling bandit (DB) algorithms enable optimal decision-making in IPE building on pairwise comparisons. However, they remain inefficient when human feedback is sparse. Existing methods address sparsity by heavily relying on parametric reward models, whose rigid assumptions are vulnerable to misspecification. In contrast, we explore an alternative perspective based on feedback augmentation, and introduce critical improvements to the model-free DB framework. Specifically, we introduce augmented confidence bounds to integrate augmented human feedback under generalized concentration properties, and analyze the multi-factored performance trade-off via regret analysis. Our prototype algorithm achieves competitive performance across several IPE benchmarks, including recommendation, multi-objective optimization, and response optimization for large language models, demonstrating the potential of our approach for provably efficient IPE in broader applications.
\end{abstract}

\begin{links}
    \link{Code}{https://github.com/COLA-Laboratory/IPEA-HF}
    % \link{Datasets}{https://aaai.org/example/datasets}
    % \link{Extended version}{https:example/extended-version}
\end{links}

\section{Introduction}
\label{sec:introduction}

\begin{figure*}[t]
    \centering
    \includegraphics[width=\linewidth]{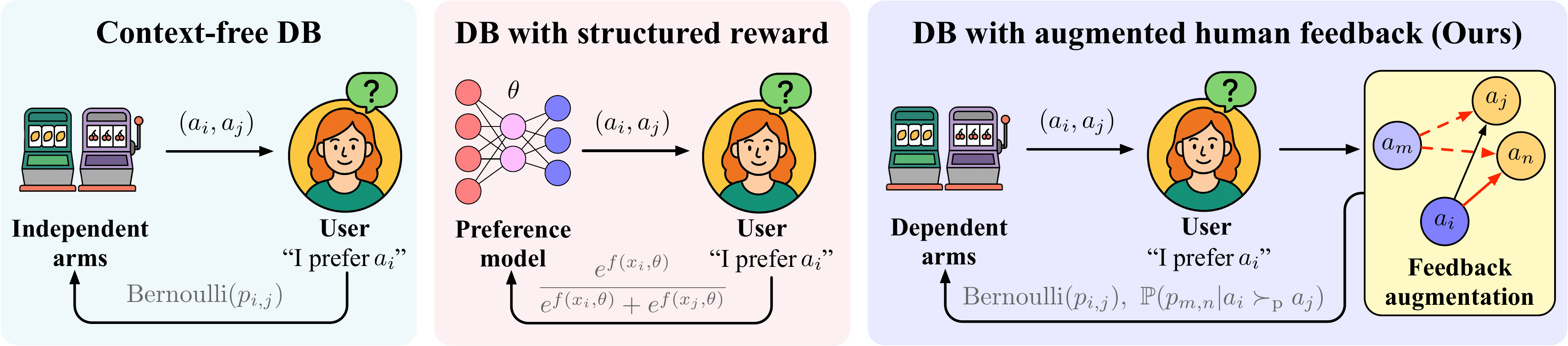}
    \caption{Comparison of three DB approaches: context-free DB, structured reward estimation, and DB with augmented feedback.}
    \label{fig:structure}
\end{figure*}

In personalization systems ranging from recommendation \cite{AustinKTS24} and multi-objective optimization \cite{HuangWL24} to large language models (LLMs) \cite{RafailovSMMEF23}, acquiring user preferences is essential but often incurs great human effort. 
Interactive preference elicitation (IPE) has the potential to substantially reduce this burden by selectively querying users through iterative strategies \cite{XiongDYWZJJZ24}. Building on pairwise comparisons, the dueling bandit (DB) framework has evolved into a strong theoretical foundation for IPE over the past decade \cite{YueBKJ12,ZoghiWMR14,Saha21}. However, even under optimal strategies, the DB framework struggles to maintain efficiency in the presence of sparse feedback, leading to waning research attention and limited practical adoption in recent years. As a result, while personalization systems continue to demand data-efficient interaction, achieving \textit{provably efficient} IPE remains a pressing challenge.

Sparse feedback in practice often reveals a deeper insight: human preferences are more than isolated pairwise comparisons. They are also shaped by cues such as contextual information and latent dependencies \cite{sun2024improving}. The independent treatment of human feedback in the DB framework may be a key source of its inefficiency. Augmenting human feedback with these cues could pave the way for a more effective DB framework that better facilitates IPE. Preliminary efforts in this direction have primarily relied on parametric preference models, most notably the Bradley–Terry (BT) model \cite{BradleyT52}. However, while such model assumptions are convenient and analytically tractable, they have been increasingly criticized for their susceptibility to model misspecification \cite{HeckelSRW16,VermaDLJL25} and suboptimal performance in the presence of non-transitive preferences \cite{MunosVCARGTGMFM24}. Alternative methods that exploit contextual information and dependencies do exist but have yet to form a widely applicable framework \cite{SuiZBY17, XiaoGSWSVCC25}. Furthermore, several fundamental questions about the DB framework with augmented human feedback remain unaddressed. In this work, we focus on the following three research questions (RQs):

\textbf{RQ1}: \textit{How can the DB framework integrate and interpret the role of augmented human feedback, while reducing reliance on rigid, predefined model assumptions?}

In the broader framework of contextual bandits \cite{LattimoreS20}, side information can be exploited through various approaches, including structured reward estimation, candidate partitioning, or similarity-based methods. Algorithms based on parametric BT models, for instance, typically fall under the reward estimation category. As discussed earlier, reward estimation risks model misspecification and struggles to accommodate non-transitive preferences. Alternatively, candidate partitioning offers favorable regret bounds that scale with the number of partition groups \cite{HuangWL24}. However, it relies on a strong assumption that candidates can be cleanly divided into distinguishable subsets, an assumption often unverifiable in real-world settings. In this work, we explore similarity-based methods to integrate augmented human feedback.
While closely related to other approaches \cite{Slivkins14}, such methods remain largely underexplored with respect to their efficiency in IPE \cite{SuiZBY17}.
Our methodology is motivated by extending the concentration results of the context-free DB framework \cite{ZoghiWMR14}. We conclude that augmenting human feedback serves as a mechanism for uncertainty calibration, offering a unifying perspective across various DB approaches. A comparison of DB approaches is presented in Figure~\ref{fig:structure}.

\textbf{RQ2}: \textit{
    Does the incorporation of augmented human feedback consistently improve the efficiency of DB algorithms, or can it sometimes introduce performance degradation?
    }

Existing studies on augmented human feedback, typically grounded in parametric or distinguishable assumptions, have shown considerable promise in improving sample efficiency. However, they often struggle to explain common pitfalls such as overfitting \cite{AzarGPMRVC24} and optimization inefficiencies \cite{RazinWSWLA25}. By discarding these predefined assumptions, we open the black box of feedback augmentation, enabling a deeper investigation into how augmented human feedback influences decision-making within the DB framework. Following the upper confidence bound (UCB) method \cite{SrinivasKKS10}, we provide theoretical analysis of sample complexity and regret bounds, revealing an explicit trade-off governed by multiple factors. These findings provide practical guidance on when to leverage, how to calibrate, and what to explore regarding human feedback augmentation, positioning the DB framework as a stronger foundation for subsequent IPE algorithm design and evaluation.

\textbf{RQ3}:  \textit{
    Can the DB framework be extended beyond pairwise comparisons to incorporate richer forms of human feedback, thereby establishing a more general foundation for IPE?
    }

Most IPE approaches, including augmented variants of the DB framework, are constrained to receiving human preferences exclusively through pairwise comparisons. However, incorporating richer forms of human feedback, such as feature-level comparisons \cite{AustinKTS24}, expert demonstrations \cite{sun2024improving}, and explanations of human choice \cite{GhazimatinPRW21}, has been shown to positively influence decision-making efficiency. As diverse feedback sources become increasingly available, the standard DB framework remains limited in its ability to directly utilize such information. 
Fortunately, our DB framework offers the potential to integrate heterogeneous forms of human feedback beyond pairwise comparisons. This is achieved by treating all feedback as a unified signals that characterize contextual similarity and dependencies among candidates. Ultimately, this would provide a more flexible foundation for IPE tasks.

We highlight the following key advances in this work:
\begin{itemize}
    \item We develop a model-free DB framework with a generalized concentration property, enabling the integration of augmented human feedback and the quantification of its influence. Our framework connects human feedback to the calibration of confidence bounds, aligning in mechanism with a broad class of DB approaches.
    \item We study the sample complexity and cumulative regret bounds in the proposed DB framework, revealing an explicit trade-off between the amount of  augmented human feedback and the strength of contextual dependencies. In addition, our analysis includes partition-based approaches \cite{HuangWL24} as a special case.
    \item We present prototype designs that incorporate a similarity-based graph structure and an auxiliary annotation process for capturing contextual dependencies. Our algorithms demonstrate competitive performance across several benchmarks, including recommendation, multi-objective optimization, and LLM response optimization. 
\end{itemize}

\section{Problem Formulation}
\label{sec:problem}

% \paragraph{IPE as dueling bandits} 
We consider an IPE task involving $K~ (K \ge 2)$ candidates, where the goal is to identify the candidate that best aligns with the user's preference, typically inferred through pairwise comparisons. Mathematically, the task can be formulated as a $K$-arms DB problem, where the set of arms is indexed by $\mathcal{A} = \{1, 2, \dots, K\}$. When contextual information is available, each arm $a_i$ is associated with a context vector $x_i\in \mathcal{X}$. In the $t$-th round, the user is asked to evaluate a pair of arms $\left(a_i, a_j\right)$, where $i,j\in\mathcal{A}$ and $i\neq j$. The user should decide, based on her/his preferences, whether $a_i$ is better, worse, or equivalent to $a_j$, denoted as $a_i\succ_\mathrm{p}a_j$, $a_i\prec_\mathrm{p}a_j$, or $a_i\simeq_\mathrm{p}a_j$. For stochastic preferences, we assume a fixed preference matrix $\mathbf{P}=[p_{i,j}]_{K\times K}$, where $p_{i,j}$ denotes the probability that arm $a_i$ is preferred over arm $a_j$ \cite{YueBKJ12}. Without loss of generality, we have $p_{i,j}+p_{j,i}=1$ and $p_{i,i}=0.5$. An arm $a_i$ is said to be superior to the $a_j$ if $p_{i,j}>0.5$. We also denote the best candidate as the winner $a_*$, defined by criteria such as the Condorcet or Copeland winners \cite{UrvoyCFN13}, among all arms.

The \textit{context-free} DB framework maintains a winning matrix $\mathbf{B}=[b_{i,j}]_{K\times K}$ to record the pairwise comparison labels, where $b_{i,j}$ denotes the number of times when arm $a_i$ is preferred over arm $a_j$. The estimated preference probability with mean $\tilde{p}_{i,j}= \frac{b_{i,j}}{b_{i,j} + b_{j,i}}$, and the upper confidence bound (UCB) $\mathbf{U}=[\tilde u_{i,j}]_{K\times K}$ and lower confidence bound (LCB) $\mathbf{L}=[\tilde l_{i,j}]_{K\times K}$ are given by \cite{ZoghiWMR14}:
\begin{equation}
        \tilde u_{i,j}=\tilde{p}_{i,j} + \sqrt{\frac{\alpha\log t}{b_{i,j}+b_{j,i}}}, ~ \tilde l_{i,j}=\tilde{p}_{i,j} -\sqrt{\frac{\alpha\log t}{b_{i,j}+b_{j,i}}},
        \label{eqn:mean-ucb-lcb}
\end{equation}
where $\alpha>0$ controls the confidence interval, and $t$ is the total number of comparisons so far. Consequently, all arms are treated independently according to their own labels. 

To encode contextual information and dependencies \cite{LattimoreS20}, a common DB approach is based on structured reward estimation. This approach assumes a structured form for ${p}_{i,j}$, such as the BT model, and a parametric reward function $f(x_i,x_j) = \theta^\top (x_i-x_j),$ with the unknown parameter $\theta$ to be determined:
\begin{equation}
    \tilde{p}_{i,j}(x_i, x_j) = \frac{1}{1 + \exp(-f(x_i,x_j))}, \quad  x_i, x_j \in \mathcal{X}.
\end{equation}
This formulation transforms the problem of identifying the best arm into one of learning the unknown parameter $\theta$. Despite its favorable sample  complexity, both the probability structure and parametric reward assumptions have been called into question \cite{HeckelSRW16, VermaDLJL25, MunosVCARGTGMFM24, AzarGPMRVC24, RazinWSWLA25}. 

Another way to leveraging contextual information is through candidate partitioning, such as clustering \cite{HuangWL24}, merging \cite{LiMRZ20}, or team grouping \cite{CohenSM21}. These methods partition the $K$ arms into $C$ subsets, where $C < K$ in order to improve sample efficiency. The partitioning process is often based on the assumption that all arms within the winner subset are strictly superior to those in other subsets \cite{JedorPL19}, a property we referred to as \textit{distinguishability}. However, candidate partitioning in real-world IPE may not always guarantee such property. 
A more robust alternative is to model dependencies through similarity \cite{SuiZBY17}. Let $s: \mathcal{X} \times \mathcal{X} \to [0, 1]$ denote a similarity function. The special case where $s\in \{0, 1\}$ corresponds to settings satisfying the distinguishable property. Nevertheless, the sample efficiency based on similarity remains unclear. We will explore this gap in the subsequent sections.

In this work, we consider the Condorcet winner \cite{YueBKJ12} for both analytical convenience and consistent benchmarking across different DB approaches. Specifically, the probability of $a_*$ satisfies $p_{*, j} > 1/2$, $\forall j \in \mathcal{A}\setminus \{*\}$. If $a_i$ and $a_j$ were chosen for comparison at time $t$, the instantaneous regret is defined as $r_t = \frac{\Delta_{i} + \Delta_{j}}{2}$, where $\Delta_{k} = p_{*, k} - \frac{1}{2}$, $\forall k \in \mathcal{A}$. Accordingly, in the context of DB, the objective of IPE is to minimize the \textit{cumulative regret up to time $T$}, defined as $R(T) = \sum_{t=1}^T r_t$. The equivalence of regret measures for parametric DB approach is demonstrated in \cite{Saha21}.

\section{Method}

\begin{algorithm}[t]
       \caption{Pseudo-code of \our}
       \label{alg:ipea-hf}
    \begin{algorithmic}% [1]
       \STATE {\bfseries Input:} Candidate number $K$, context set $\mathcal{X}$, $\alpha>0$
       \STATE {\bfseries Init:}  $\mathbf{B} = [0]_{K\times K}$, graph $\mathcal{G}(\mathcal{X}, K)$, $ W=[0]$.
       \REPEAT
       \STATE $\mathbf{U}, \mathbf{L} \leftarrow \textcolor{blue}{\texttt{AugConfidanceBound}}(\mathbf{B},\mathcal{G}, W, \alpha)$ 
       \STATE Select pairs $(a_i, a_j) \leftarrow \textcolor{blue}{\texttt{DuelingBanditAlgo}}(\mathbf{U}, \mathbf{L})$
       \STATE Observe $a_{\text{win}}$, $a_{\text{lose}}$ and update $b_{\text{win}, \text{lose}} \leftarrow b_{\text{win}, \text{lose}} + 1$
       \STATE $W \leftarrow \textcolor{blue}{\texttt{DependencyExtract}}(\mathcal{X}, \mathcal{G}, W)$
       \STATE $\mathcal{G}  \leftarrow \textcolor{blue}{\texttt{FeedbackAug}}(a_{\text{win}},a_{\text{lose}}, W, \mathcal{G})$
       \UNTIL{IPE task finished or budget exhausted}
    \end{algorithmic}
\end{algorithm}

In this section, we delineate the DB framework for IPE with augmented human feedback, referred to as \our. The overall structure is outlined in Algorithm \ref{alg:ipea-hf}, which compromises four key components. It begins by introducing augmented confidence bounds that incorporate augmented human feedback in a model-free setting. This is followed by pair selection criteria, which depend on the specific DB algorithm employed, such as the relative upper confidence bound (RUCB) \cite{ZoghiWMR14} or double Thompson sampling (DTS) \cite{WuL16}. Upon observing a user response, \our\ augments the feedback based on contextual similarity and extracted dependencies through computational algorithm designs and additional annotations.

\subsection{\texttt{AugConfidanceBound}: Integrating Confidence Bounds with Augmented Feedback}

The confidence bounds in the DB framework play a pivotal role in determining which pairs are selected for comparison, and ultimately shape the overall query process. As formalized in equation~\eqref{eqn:mean-ucb-lcb}, bounds $\tilde u_{i,j}$ and $\tilde l_{i,j}$ in the context-free DB approach rely solely on \textit{direct} observations from comparisons between the pair $(a_i, a_j)$, specifically the counts $b_{i,j}$ and $b_{j,i}$.  In contrast, we explore whether incorporating additional \textit{related} observations from augmented human feedback can enhance the estimation of confidence bounds. 
% In particular, we investigate whether feedback derived from other pairwise comparisons can complement the direct comparison-based updates. 
For $i, j \in \mathcal{A}$, let $n^d_{i,j}(t)= b_{i,j} + b_{j,i}$ denote the number of direct comparisons between $a_i$ and $a_j$ up to time $t$, and let $n^r_{i,j}(t)$ denote the number of related observations inferred from other pairwise comparisons up to time $t$. We then define the total observation count as $n_{i,j}(t) = n^d_{i,j}(t) + n^r_{i,j}(t)$. The augmented mean becomes $\hat p_{i,j} = \frac{1}{\eta n_{i,j}(t) } \left(b_{i,j} + \sum_{k=1}^{n^r_{i,j}} X_{i,j}^k\right)$, and the UCB and LCB take the following form:
\begin{equation}
    \hat u_{i,j} = \hat p_{i,j} +  \frac{1}{\eta}\sqrt{\frac{\alpha \ln t}{n_{i,j}(t)}}, ~\hat l_{i,j} = \hat p_{i,j} -  \frac{1}{\eta}\sqrt{\frac{\alpha \ln t}{n_{i,j}(t)}}.
        \label{eqn:augmented-ucb}
\end{equation}
where $\eta = \left(n^d_{i,j} + \sum_{k=1}^{n^r_{i,j}} w^k_{i,j}\right) / {n_{i,j}}$ with $w^k_{i,j} \in [0,1]$ to be determined, and $X_{i,j}^k$ denotes a random variable in $[0, 1]$ determined by related observations. By definition, without augmented feedback, i.e., $n^r_{i,j} = 0$, the augmented confidence bounds reduce to model-free ones in equation \eqref{eqn:mean-ucb-lcb}. 

\begin{figure*}[t]
    \centering
    \includegraphics[width=\linewidth]{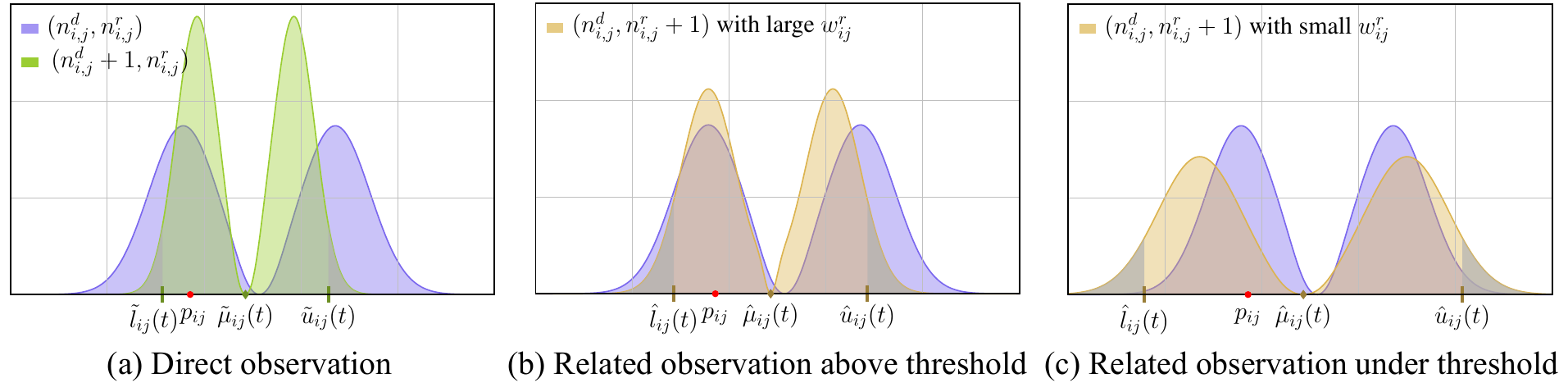}
    \caption{Comparison of confidence intervals adding a direct observation or a related observation. The curve shows the derivative of the bound in equation~\eqref{eqn:thm1-concentration}, where the shaded area is the probability that the confidence interval fails to contain $p_{i,j}$. % Refer to \cite{ZoghiWMR14} for details.
    }
    \label{fig:intevals}
\end{figure*}

\paragraph{Concentration property} 
We provide the following concentration property of the augmented confidence bounds as a generalized result from the context-free DB framework \cite{ZoghiWMR14,WuL16}. 

\begin{theorem}
    \label{thm:concentration}
    Assume $X^k_{i,j} \sim \mathrm{Bernoulli}( w^k_{i,j} p_{i,j})$ and let $C(\delta) = \left(\frac{(4 \alpha - 1) K^2}{(2 \alpha - 1) \delta}\right)^{\frac{1}{2\alpha - 1}}$. Given the preference matrix $\mathbf{P}$ with $K$ arms, then, for any $\alpha > 0.5$ and $\delta \in (0, 1)$, we have:
    \begin{equation}
    \label{eqn:thm1-concentration}
        {P}\left( \forall t > C(\delta),~ i, j\in \mathcal{A}, ~ p_{i,j} \in \left[ \hat l_{i,j}(t),~ \hat u_{i, j}(t) \right] \right) > 1 - \delta.
    \end{equation}
\end{theorem}
The proof is given in Appendix~\ref{sec:appendix_theoretical_concentration}. This result suggests that additional observations can be leveraged to inform the estimation of confidence bounds in conjunction with the latent dependency weight $w^k_{i,j}$. We will later discuss how to obtain the weights in Section~\ref{sec:method-computationaldesign}. 

We now study the property of concentration. 
Consider the case where $n_{i,j}^r(t) = 0$. Given a direct observation, the confidence interval shrinks from $\sqrt{\alpha \ln t / n^d_{{i,j}}(t)}$ to $\sqrt{\alpha \ln t / (n^d_{{i,j}}(t) + 1)}$. The confidence intervals progressively narrow as the number of direct comparisons increases \cite{ZoghiWMR14}. By contrast, for a related observation, the confidence bound takes the form $\frac{1}{\eta}\sqrt{\alpha \ln t / (n^d_{{i,j}}(t) + 1)}$. Notably, the weighting term $w^k_{i,j}$ influences the behavior of the confidence bound. When $w_{i,j}^r = 1$, we have $\eta=1$, and the confidence bound shrinks in the same way as with a direct observation. Conversely, when $w_{i,j}^r = 0$, the bound becomes $\sqrt{\alpha (n^d_{{i,j}}(t) + 1) \ln t / n^{d~2}_{{i,j}}(t)}$, which is strictly larger than the original bound $ \sqrt{\alpha \ln t / n^{d}_{{i,j}}(t)}$. Therefore, both direct and relative observations contribute to the \textit{uncertainty calibration} for the confidence bounds, with the effect of relative observations depending on the strengths of dependencies.

\paragraph{Calibration threshold} When estimating $p_{i,j}$, a smaller confidence interval is desirable for making more informed decisions. We can quantify the impact of augmented feedback by the  ratio of confidence intervals with and without a relative observation:
\begin{align}
   \frac{\hat u_{i,j} - \hat l_{i,j}}{\tilde u_{i, j} - \tilde l_{i, j} } = & \frac{1}{\eta}\sqrt{\frac{\alpha \ln t}{n^d_{i,j}(t)+ 1}} /  \sqrt{\frac{\alpha \ln t}{ n^d_{i,j}(t)}}   \nonumber \\
   \in & \left[\sqrt{1 - \frac{1}{n^d_{i,j}(t) + 1}},~ \sqrt{1 + \frac{1 }{n^d_{i,j}(t)}}~\right].
   \label{eqn:ratio}
\end{align}
We can derive a threshold condition for identifying \textit{good} feedback augmentation that facilitates uncertainty calibration:
\begin{equation}
   w_{i,j}^r > \eta  n_{i, j}(t) \left(\sqrt{1 + \frac{1}{n_{i.j}(t)}} - 1 \right).
   \label{eqn:cali_threshold}
\end{equation}
This condition holds for all $n_r \ge 0$. The detailed derivation is provided in Appendix~\ref{sec:appendix_theoretical_threshold}.
The result indicates that the contribution of a relative observation increases monotonically with the strength of $w_{i,j}^r$, becomes equivalent to a direct observation when $w_{i,j}^r = 1$, and diminishes as the number of direct observations increases, as illustrated in Figure~\ref{fig:intevals}.

\paragraph{Connections to other DB approaches} 
The augmented confidence bounds naturally encompass partition-based approaches as a special case. Specifically, for all arms within a distinguishable group, related observations correspond to complete dependencies, i.e., $w_{i, j}^r = 1$. By considering only related observations within the same group, the comparison between individual arms effectively reduces to a comparison between groups, mirroring the operations adopted in partition-based approaches. For DB approaches with structured reward, they maintain a pairwise score of the form $u_f(x_i, x_j) = \hat \theta (x_i - x_j) + \sigma \Vert x_i - x_j \Vert_{V^{-1}}$ where $\hat \theta$ is the estimated parameter, $\sigma>0$, and $V$ is the sum of outer products of compared pairs \cite{VermaDLJL25, Saha21}. The second term of $u_f(x_i, x_j)$ represents the confidence bound expressed as a Mahalanobis norm \cite{2018ReprintOM}. For a pair $(a_i, a_j)$, more frequent comparisons, i.e., more direct observations in our setting, assign smaller weight to the direction of $ x_i - x_j$. Given a related observation that is close in distance to $(a_i, a_j)$, representing stronger dependencies, the weight is further reduced along that direction, indicating a narrowed uncertainty calibration for the score of $(a_i, a_j)$. This mechanism that lowers the direction weight in the presence of high-correlated observations aligns with our proposed method. However, the confidence bound in structured reward approaches remains heuristic and lacks formal guarantees on the concentration property (i.e., $p_{i,j}$ may not be included within the bound with high probability as $t$ increases beyond a certain threshold), making it prone to overfitting due to model misspecification.

\begin{questionbox}
    \textbf{Insights on RQ1}:  \textit{
    By leveraging the generalized concentration property, our method fundamentally strengthens the model-free DB framework by incorporating related observations into the estimation of winning probabilities and confidence bounds. We identify the key conditions under which contextual correlations enhance uncertainty calibration and establish clear connections to structured and partition-based approaches, highlighting the broader applicability.
    }
\end{questionbox}

\begin{figure*}
  \centering
  \includegraphics[width=\textwidth]{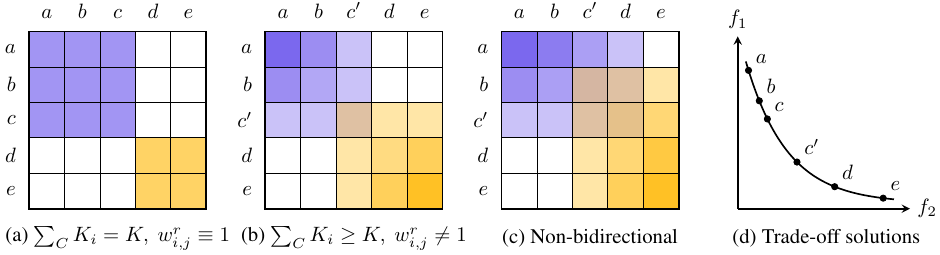}
  \caption{Illustrative cases for bidirectional dependency. (a) Candidate partitioning. (b) Dependent arms with symmetric correlations. (c) General case without bidirectional dependency. (d) Two-objective trade-off solutions whose dependencies are determined by their distances. Given $[a,b,c,d,e]$, the candidates can be safely grouped into two subsets. When $c$ shifts to $c^\prime$, the dependencies weaken, making soft clustering a more suitable choice.}
  \label{fig:cluster}
\end{figure*}

\subsection{\texttt{DuelingBanditAlgo}: Decision-Making with Augmented Feedback}
The estimation of winning probabilities and their confidence bounds is integrated into DB decision-making algorithms to strike a balance between \textit{exploitation} and \textit{exploration}. These algorithms select candidate pairs through a two-stage process:
\begin{enumerate}%[label=\textbf{Stage \arabic*}]
    \item (Exploitation) Select the most promising arm as the first candidate. This is achieved by identifying the arm whose UCBs outperform the largest number of other arms.
    \item  (Exploration) Select the most competitive arm against the first candidate. This aims to identify the arm likely to beat the first selected arm according to their UCB and LCB.
\end{enumerate}
Existing DB algorithms differ in their selection mechanism, including both deterministic and stochastic strategies, such as RUCB \cite{ZoghiWMR14} and DTS \cite{WuL16}. Our augmented confidence bounds can be seamlessly integrated into both types. We propose two DB algorithms: \texttt{IPEA-RUCB} and \texttt{IPEA-DTS} integrated with augmented confidence bounds, with detailed descriptions in Appendix~\ref{sec:appendix_algorithm}.

\paragraph{Sample Efficiency} Following the analysis in \cite{ZoghiWMR14}, we investigate the RUCB variant integrated with augmented confidence bounds. Since our framework incorporates both direct observations (real samples) and related observations (virtual samples), the following result establishes a high-probability bound on the total sample complexity of each suboptimal arm pair.

\begin{theorem}
    \label{thm:sample_number}
    Given the setup in Theorem~\ref{thm:concentration}, and let $D_{i,j}^w = \frac{4\alpha}{\min_r w_{i,j}^{r~2}\min\{\Delta_i^2, \Delta_j^2\} }$. For the \texttt{IPEA-RUCB} algorithm and any suboptimal pair $(a_i,a_j)\ne (a_*,a_*)$, $n_{i,j}(t)$ satisfies
    \begin{equation}
    \label{eqn:thm2-total_comparison_number}
        P(\exists t,~ i,j \in \mathcal{A},~ n_{i,j}(t)>C(\delta)\lor D_{i,j}^w\ln t) < \delta.
    \end{equation}
\end{theorem}
The proof is given in Appendix~\ref{sec:appendix_theoretical_compare_number}. Accordingly, the multi-factored sample complexity reveals a fundamental \textbf{trade-off} in integrating augmented feedback. On the positive side, for a pair $(a_i, a_j)$ that utilizes related observations of $(a_m, a_n)$, it holds that $n^d_{i,j}(t) + n^d_{m,n}(t) \leq n_{i,j}(t)$. That is, the number of direct observations of one pair that provide augmented feedback to other pairs is already included in the total sample count of the latter. If more pairs are related, the total sampling count satisfies $n^d_{i,j}(t) + \sum_{m,n} n^d_{m,n}(t) \leq n_{i,j}(t)$. This indicates that incorporating related observations 
% redistributes existing sample contributions, with the 
has the potential to improve overall sample efficiency. However, for less dependent observations, e.g., when $\min_r w^r_{i,j} \to 0$, $D^w_{i,j}$ can increase significantly, introducing a large coefficient into the overall sample complexity. According to this trade-off, augmented feedback should be integrated selectively, prioritizing observations with strong contextual dependencies.

\paragraph{Regret Analysis} To conduct regret analysis, we assume bidirectional dependency: if a pair $(a_i, a_j)$ utilizes related observations from another pair $(a_m, a_n)$, the latter also utilizes the observations from $(a_i, a_j)$. A direct consequence of this assumption is the soft-clustering, grouping candidates into $C$ subsets. Notably, existing methods based on distinguishability \cite{LiMRZ20,CohenSM21} or arm dependencies \cite{SuiZBY17} satisfy this assumption. Furthermore, our method generalizes beyond these settings, as illustrated by the diverse cases in Figure~\ref{fig:cluster}.

\begin{theorem}
    \label{thm:regret}
    Follow the setup in Theorems~\ref{thm:concentration} and~\ref{thm:sample_number} and assume bidirectional dependencies. If we have $C$ soft clusters and each cluster contains $K_i$ candidates, the cumulated regret bound of \texttt{IPEA-RUCB} is $ \mathcal{O}\left( \frac{1}{\min_{i,j,r}{w_{i,j}^r}}\hat K^2 \log T\right)$, where $\hat K = \max\{C, K_1, \dots, K_C\}$.
\end{theorem}
The proof is provided in Appendix~\ref{thm:appendix_theoretical_regret}. This result subsumes the partition-based DB approach as a special case. For example, the cluster-based DB method achieves a regret bound of $\mathcal{O}\left(C^2 \log T\right)$ \cite{HuangWL24}, where the analysis is conducted at the cluster level under the assumption of distinguishability ($w_{i,j}^r \equiv 1 $). In contrast, our result operates at the candidate level and explicitly accounts for varying degree of dependency, such that the candidate count in each subset is also reflected in the effective parameter $\hat K$. In conclusion, our regret bound generalizes the partition-based approach \cite{HuangWL24} and offers promising support for DB algorithms that incorporate correlations or dependent arms \cite{SuiZBY17}.
\begin{questionbox}
    \textbf{Insights on RQ2}:  \textit{
    Our analysis of sample complexity and regret bounds reveals a fundamental trade-off in incorporating augmented human feedback into DB algorithms. Theoretical results consistently show that related observations with strong dependencies can substantially reduce the number of required interactions, thereby improving efficiency. This finding aligns with existing conclusions and offers generalized support to model-free, dependency-aware DB approaches.
    }
\end{questionbox}

\subsection{\texttt{DependencyExtract \& FeedbackAug}: Computational Design}
\label{sec:method-computationaldesign}

Our algorithm consists of two key components: extracting dependencies and augmenting human feedback. Our analysis (Theorem~\ref{thm:concentration}) characterizes latent dependencies as the joint distribution of direct and indirect observations, formalized as $\mathrm{Bernoulli}( w^k_{i,j} p_{i,j})$. The weighting term $w^k_{i,j}$ captures the conditional dependency between pairs.

\begin{figure}
    \includegraphics[width=0.5\textwidth]{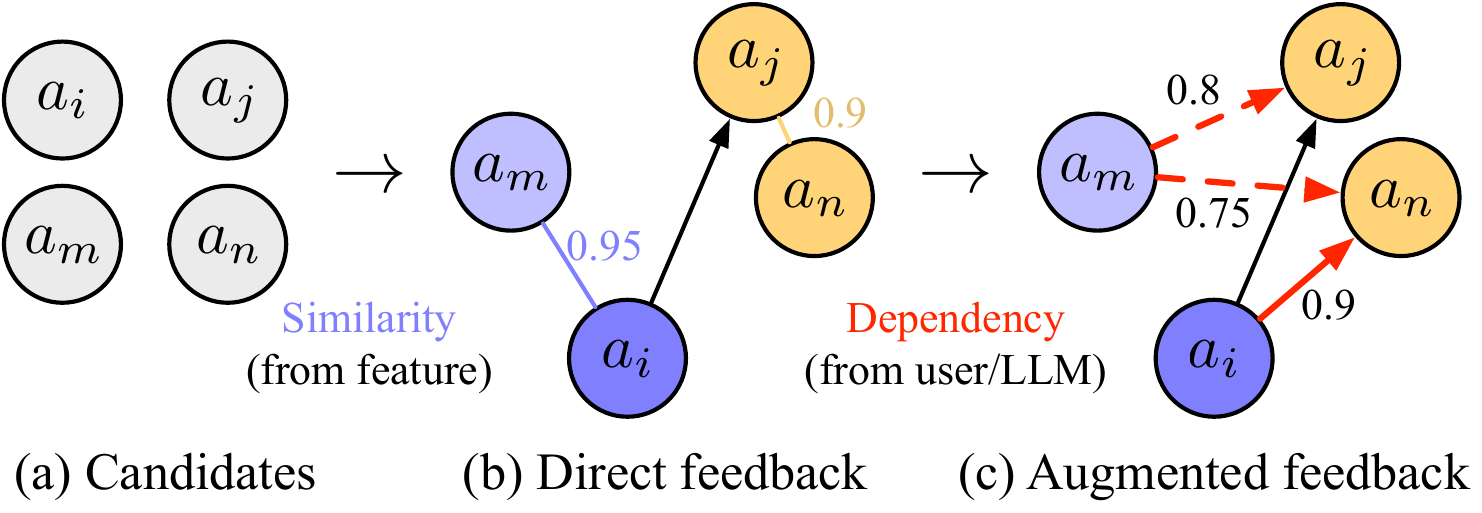}
    \caption{Illustration of computational design of dependency extraction and feedback augmentation.}
    \label{fig:augmentation}
\end{figure}

To estimate $w_{i,j}^r$, we follow a two-step procedure, as illustrated in Figure~\ref{fig:augmentation}. First, we construct a prior structural skeleton by building a similarity graph over the candidates, where the edges are determined by similarity distances in the context space $\mathcal{X}$. We filter out dissimilar pairs using graph partitioning or soft clustering, thereby concentrating on the most meaningful dependencies. As a result, each candidate is associated with one or more groups, depending on the employed partitioning method. At each round of pairwise comparisons, we observe human feedback on a pair, e.g., $a_i$ is preferred over $a_j$. Then, we should determine the dependencies. Through  additional annotations, we selectively query user or LLMs to annotate conditional dependencies over local pairwise relations, and incorporate related observations with high dependency scores into the DB loop. Specifically, we iterate over and combine all pairs from the groups associated with $a_i$ and $a_j$. When $a_i$ and $a_j$ belong to the same group, no feedback augmentation is performed.

While the annotation process for dependency estimation provides valuable signals for feedback augmentation, it may introduce errors, particularly when relying on simulated annotators such as LLMs. The robustness of our framework stems from the fact that the influence of a related observation is strictly weaker than that of a direct observation, and further diminishes as the number of direct observations increases (see equation~\eqref{eqn:ratio}). In addition, the impact of biased estimation in DB has been analyzed in \cite{yiKL24}. As interactions accumulate, the estimated dependencies are either refined or discarded, enabling the algorithm to recover from initial noisy annotations over time.

\paragraph{Richer Forms of Human Feedback} In the modelling of conditional dependencies, multiple forms of feedback can be leveraged. Beyond pairwise comparisons and contextual similarities, our method also incorporates reasoning signals derived from LLMs or domain experts on latent contextual dependencies. Additional types of data, such as human explanations and rationales \cite{GhazimatinPRW21}, can also be readily integrated, all contributing to the discovery and refinement of informative conditional dependencies. This enables the incorporation of richer feedback signals within a principled framework, supported by the theoretical foundations of our DB framework. As more dependency data becomes available, probabilistic graphical models can be employed to capture joint distributions and conditional structures, thereby further denoising sparse human feedback and estimating missing relationships.

\section{Empirical Study}
\label{sec:empirical_study}
% \subsection{Experimental Settings}
We evaluate our algorithms (\texttt{IPEA-RUCB} and \texttt{IPEA-DTS}) across a diverse set of benchmarks:
% , including item recommendation, multi-objective optimization, and exploration of LLMs:
\begin{itemize}
    \item \textbf{Item Recommendation.} We consider the Sushi dataset containing full rankings over $10$ types of sushi collected from $5{,}000$ customer orders \cite{KamishimaA10}, and a Car Preference dataset with full rankings over $10$ cars provided by $60$ U.S. users via Amazon Mechanical Turk \cite{AbbasnejadSBP13}. 
    
    \item \textbf{Multi-Objective Optimization.} We adopt the \textit{a posteriori} setting of preference-based evolutionary multi-objective optimization (PBEMO) \cite{HuangWL24}, considering the celebrated synthetic test problems (DTLZ2 and DTLZ7) with different landscapes \cite{DebTLZ05}.
    
    \item \textbf{LLM Response Optimization}. We follow the experimental settings in \cite{VermaDLJL25,DwaracherlaAHR24} to conduct \textit{active exploration} given a pool of responses for each prompt sampled from Anthropic Helpfulness and Harmlessness (H-H) dataset \cite{BaiJNACDD22}. %We leverage the fine-tuned GPT model to simulate human feedback \cite{radford2019language} and apply the BT model to compute winning probabilities.
\end{itemize}
We consider five state-of-the-art DB methods spanning context-free and parametric frameworks. 
% Partition-based approaches are excluded, as they target the best subset rather than a single winner:
\begin{itemize}
    \item \textbf{Context-free algorithms.} We include two prominent DB algorithms: Relative Upper Confidence Bound (\texttt{RUCB}) \cite{ZoghiWMR14} and Double Thompson Sampling (\texttt{DTS}) \cite{WuL16}.
    \item \textbf{Parametric algorithms.} We select three state-of-the-art methods based on the BT models: Maximum Informative Pair (\texttt{MaxInP}) \cite{Saha21}, \texttt{COLSTIM} \cite{BengsSH22}, and \texttt{VACDB} \cite{Di0WZFG24}.
\end{itemize}

Detailed experimental settings are given in Appendix \ref{sec:appendix_exp}. 
% For each task, we run each algorithm $10$ times. 

% \subsection{Experimental Results}
\begin{figure}[t]
    \centering
    \includegraphics[width=\linewidth]{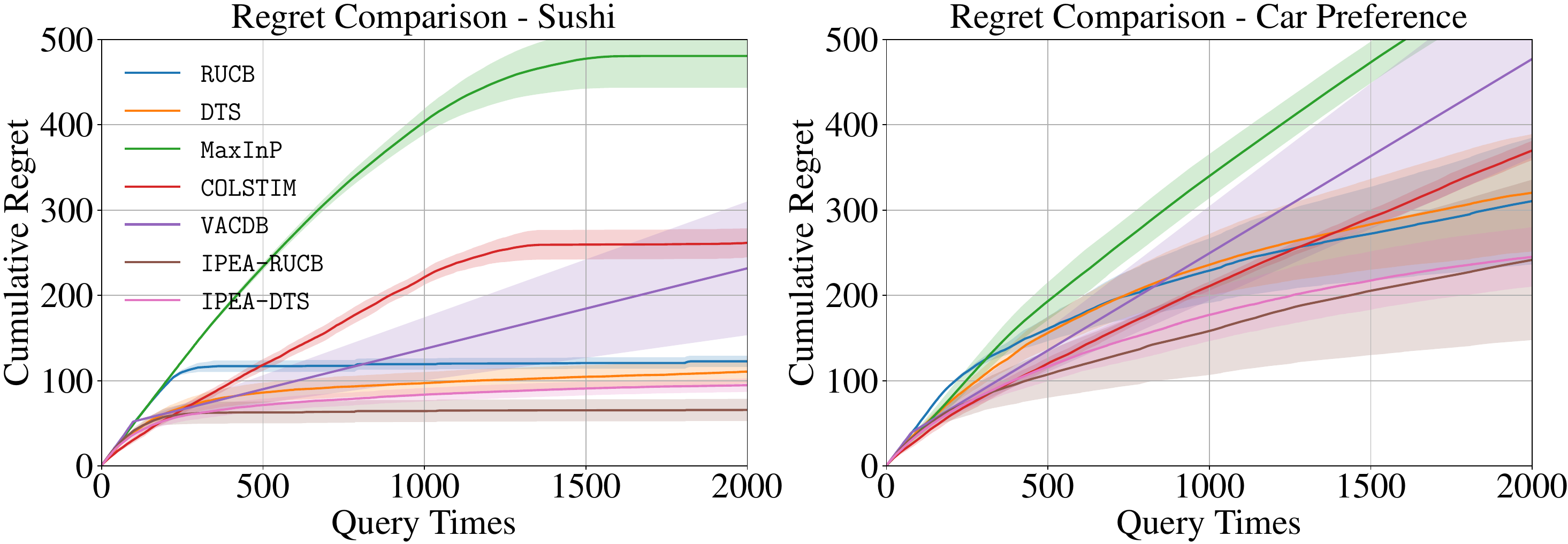}
    \caption{Regret trajectories on Sushi and Car Preference.}
    \label{fig:rec_result}
\end{figure}
\paragraph{Item Recommendation} We simulate 2,000 interaction rounds to reflect the aggregated preferences of a user group. The cumulative regret of all algorithms is shown in Figure~\ref{fig:rec_result}. Under a moderate number of candidate items, context-free DB algorithms consistently outperform parametric baselines. Building on this foundation, our proposed algorithms achieve further improvements in sample efficiency. The inferior performance of parametric methods is largely due to model misspecification, which is exacerbated when handling real-world, mixed-type feature spaces. Moreover, some parametric algorithms suffer from premature convergence. For example, \texttt{VACDB} exhibits a steadily increasing regret curve, indicating that it fails to maintain effective exploration over time.

\begin{figure}
    \centering
    \includegraphics[width=\linewidth]{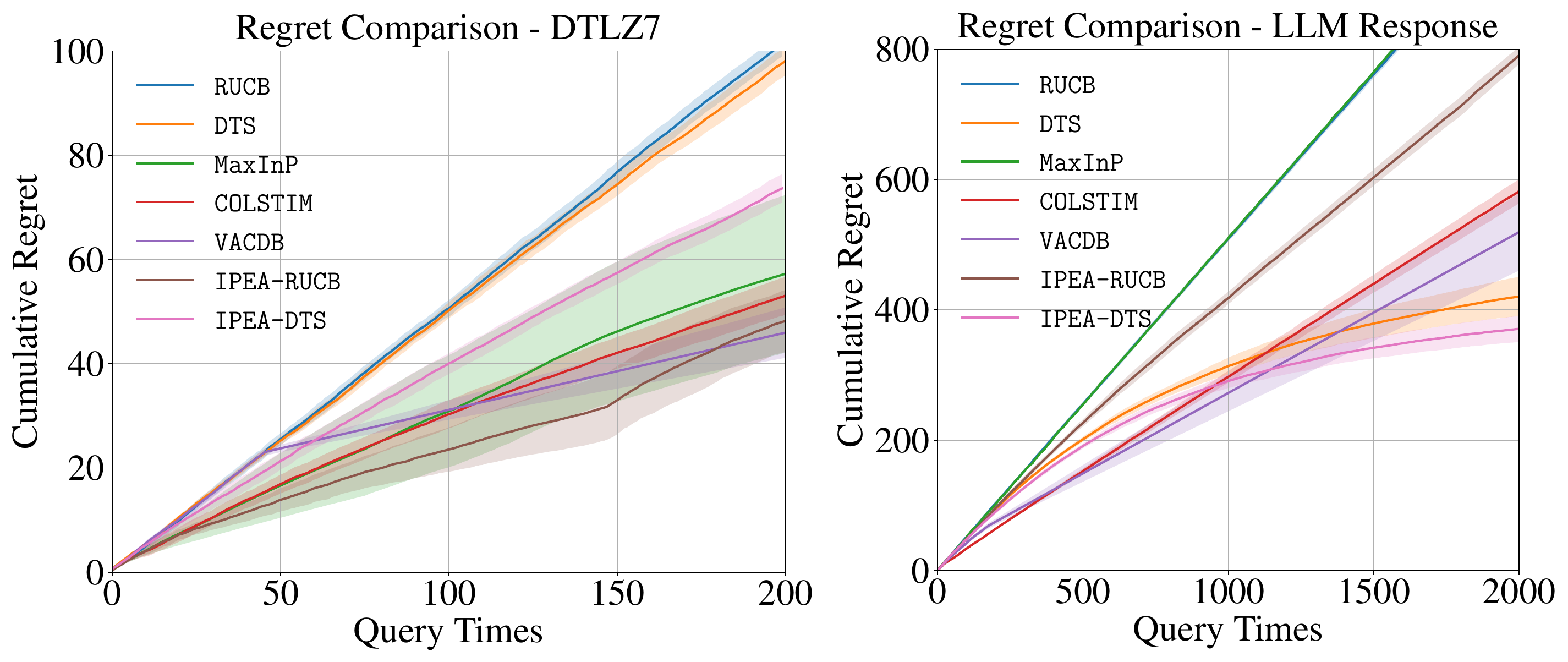}
    \caption{Regret on DTLZ7 and LLM Exploration.}
    \label{fig:tranj_dtlz_resp}
\end{figure}

\begin{figure}
    \centering
    \includegraphics[width=1\linewidth]{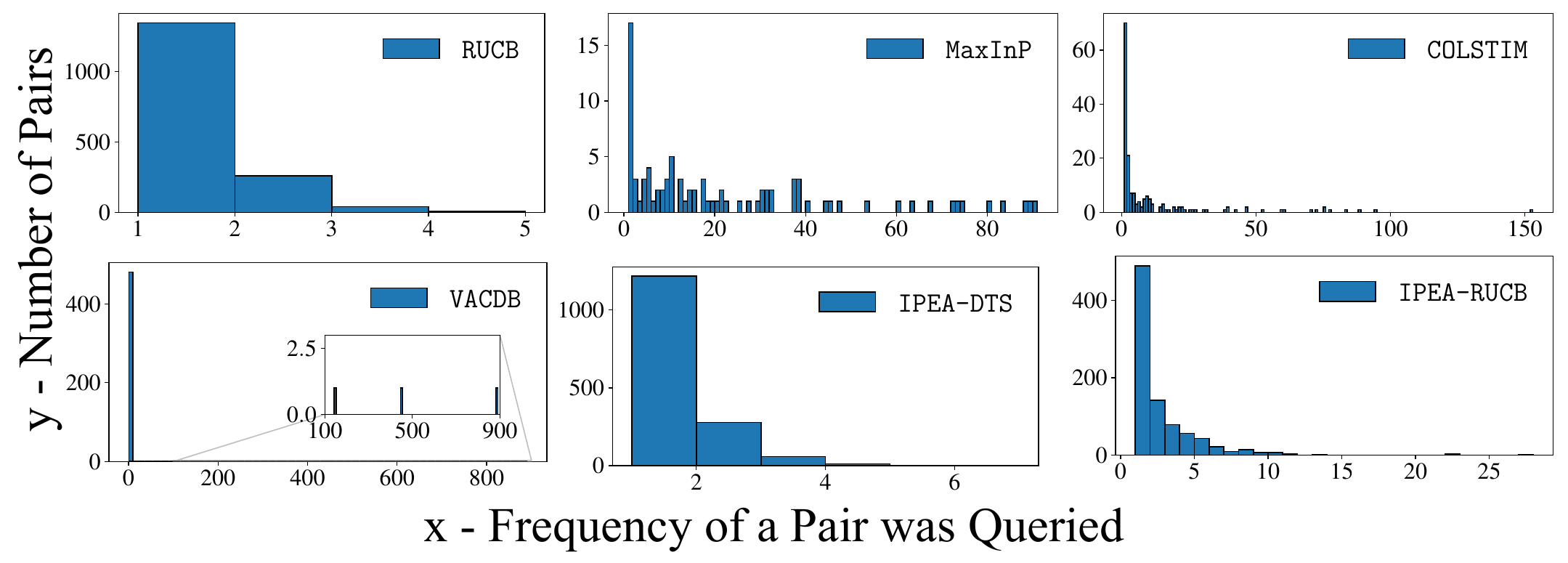}
    \caption{Analysis of Query frequency on the DTLZ7 task. In the histogram, $(2, 1000)$ denotes 1,000 unique pairs each queried twice. A broader spread at low frequencies indicates greater exploration.}
    \label{fig:frequency}
\end{figure}

\paragraph{Multi-Objective Optimization}
We simulate 200 interaction rounds to evaluate algorithm performance under sparse human feedback. The cumulative regret trajectories are shown in the left panel of Figure~\ref{fig:tranj_dtlz_resp}. As expected, the standard DB algorithms \texttt{RUCB} and \texttt{DTS} exhibit poor efficiency when facing a large candidate set ($100^2$ pairs) and a limited interaction budget. Although parametric DB algorithms demonstrate relatively better performance, their limitations are evident. Regarding the query frequency as given in Figure~\ref{fig:frequency}, both \texttt{COLSTIM} and \texttt{VACDB} repeatedly select a small subset of pairs, reflecting strong exploitation and a lack of sufficient exploration. In contrast, our proposed algorithms achieve a better trade-off between regret minimization and query diversity. Notably, \texttt{IPEA-RUCB} performs more favorably than \texttt{IPEA-DTS} under these conditions, suggesting that deterministic strategies may be more robust to sparse feedback.

\paragraph{Response Optimization}
The regret trajectories are shown in the right panel of Figure~\ref{fig:tranj_dtlz_resp}. Most algorithms exhibit a degree of inefficiency, with the exception of \texttt{DTS} and \texttt{IPEA-DTS}, both employ a stochastic strategy. This aligns with the empirical findings of \cite{DwaracherlaAHR24}, where DTS was shown to be the most effective algorithm for active exploration for LLMs. Notably, by integrating augmented human feedback, our algorithms consistently outperform their context-free counterparts, \texttt{RUCB} and \texttt{DTS}. In contrast, algorithms based on parametric models require significantly more computational resource, due to the high dimensionality of the embedded feature space (768 dimensions from the MPNet \cite{Song0QLL20}). As a result, without a dedicated context-aware mechanism \cite{VermaDLJL25}, the effectiveness using parametric DB for response optimization remains questionable.

In response optimization, our algorithms leverage not only related observations from the same prompt but also inter-prompt observations with high contextual similarity and dependency. This pilot demonstration shows the effectiveness of feedback augmentation beyond pairwise comparisons.

\begin{questionbox}
    \textbf{Insights on RQ3}:  \textit{
    By treating human feedback as unified signals characterizing contextual similarity and dependencies, our prototype computation designs are readily extensible to incorporate richer forms of feedback. In our experiments, we integrate contextual similarity, LLM-powered dependency annotations, and inter-prompt observations, each serving as an additional form of human feedback. This expands the applicability of the DB framework, offering a promising pathway for adapting to diverse feedback modalities in future IPE tasks.
    }
\end{questionbox}

\section{Conclusions}
We introduced a novel DB framework that integrates augmented human feedback to enable provably efficient IPE. By analyzing the concentration properties and performance trade-offs, we demonstrated the effectiveness of feedback augmentation and established connections to a broad class of existing DB approaches. In addition, we proposed prototype computational designs that incorporate richer forms of human feedback, and showed competitive performance across diverse benchmarks. This work offers a principled, extensible foundation for the development of personalization systems.

\section{Acknowledgments}
We sincerely thank all the reviewers for their encouraging and constructive feedback. This work was supported by the UKRI Future Leaders Fellowship under Grant MR/S017062/1 and MR/X011135/1; in part by NSFC under Grant 62376056 and 62076056; in part by the Royal Society Faraday Discovery Fellowship (FDF/S2/251014), BBSRC Transformative Research Technologies (UKRI1875), Royal Society International Exchanges Award (IES/R3/243136), Kan Tong Po Fellowship (KTP/R1/231017); and the Amazon Research Award and Alan Turing Fellowship.

\bibliography{aaai2026}

\onecolumn
\onecolumn
% \section*{Appendix for "Preference is More Than Comparisons: Rethinking Dueling Bandits with Augmented Human Feedback"}

\section{Related Work}
\label{sec:related_work}
\subsection{Dueling Bandits}
Dueling bandits (DB), which leverage duel feedback for online decision-making, were first introduced by \cite{YueBKJ12} and have since become a core paradigm in IPE. Depending on the availability of additional information, DB algorithms are generally categorized into two classes.
\paragraph{Stochastic Dueling Bandits}
In stochastic DB, each pair of arms is associated with an unknown winning probability. Early work relied on strong stochastic transitivity assumptions \cite{YueBKJ12}, while subsequent research has shifted toward more general definition of winners. The Condorcet winner, as used in our work, was introduced in \cite{UrvoyCFN13}. Building on this, the RUCB algorithm was proposed in \cite{ZoghiWMR14}, establishing insightful theoretical foundations for the DB framework. Another important type of winner is the Copeland winner \cite{ZoghiKWR15}, which generalizes the Condorcet assumption and underpins the analysis of the DTS algorithm \cite{WuL16}. The Borda winner, as introduced in \cite{JamiesonKDN15}, enables a theoretically efficient DB algorithm under sparse feedback. Different from our method, it requires additional structural assumptions on the preference matrix to ensure sample efficiency \cite{JamiesonKDN15}. In this work, we adopt the Copeland winner assumption due to its compatibility with UCB-based decision-making and its widespread use in prior studies (see the regret equivalence in \cite{Saha21}). Extensions of stochastic DB algorithms to more complex scenarios including adversarial preferences \cite{SahaG22} and delayed feedback \cite{yiKL24} have also been proposed, offering potential robustness verification of our methods.

\paragraph{Contextual Dueling Bandits}
In our work, context provides side information that encodes similarity or structural relationships among candidates. This notion aligns with broader research in preference optimization, including reinforcement learning with human feedback \cite{RafailovSMMEF23} and preferential optimization \cite{Gonzalez17a}. A different notion of context is considered in \cite{DudikHSSZ15}, where the context is chosen by nature and reflects the dynamic structure of the preference matrix across time. To leverage the context-based side information, existing methods typically fall into three categories:
\begin{itemize}
	\item \textbf{Structured reward estimation.} By predefining a shared reward structure originating from linear bandits \cite{LattimoreS20}, the objective in IPE shifts from identifying the most preferred candidate to estimating the  underlying utility function. Algorithms and variants that generalize to subset-wise comparisons were proposed in \cite{Saha21}. Following on, more flexible parametric models are studied in \cite{BengsSH22,Di0WZFG24}. Most recently, to mitigate the risks of model misspecification, neural estimator has been introduced in \cite{VermaDLJL25}. Most of these methods adopt deterministic strategies, while random strategy such as Thompson sampling has been recently integrated as a promising alternatives for linear contextual DB \cite{LiZG24}. A key limitation of this line of work is the inability to quantify the impact of misleading model assumptions. In contrast, our method explicitly captures both the potential benefits and risks of feedback augmentation, without requiring a predefined parametric structure.
	\item \textbf{Candidate partitioning.} To reduce search and sample complexity, early work leveraged contextual similarity to partition candidates into subsets \cite{Slivkins14}. This transforms the objective from identifying the single most preferred candidate to locating coarse-grained regions of interest \cite{HuangWL24}. In DB literature, partitioning strategies include clustering \cite{HuangWL24}, merging \cite{LiMRZ20}, and grouping \cite{CohenSM21}. These approaches often fail to capture fine-grained preferences or more complex contextual relationships. An exception is MergeDTS \cite{LiMRZ20}, whose sample complexity remains comparable to non-merged alternatives. Our framework generalizes partition-based DB methods as a special case and retains the ability to identify the best single candidate, enabling finer preference resolution when needed.
	\item \textbf{Similarity and dependency.} There are limited studies on similarity functions \cite{LattimoreS20} and dependent arms \cite{SuiZBY17}. As suggested by our results, similarity alone is often insufficient to fully capture the relational structure among candidates. Motivated by this limitation, our approach shifts focus from raw similarity to more expressive notions of dependency, which we estimate through annotations derived from richer forms of human feedback. This allows us to capture more nuanced, asymmetric, and context-sensitive relationships, ultimately enabling more effective augmentation of human preferences.
\end{itemize}

\subsection{Augmentation and Richer Forms of Human Feedback}

In addition to the DB framework, several algorithms for IPE and general preference optimization have explored the integration of richer forms of feedback through augmentation. When candidate items are associated with feature-level keys, both the items and their attributes can serve as contextual information. The relationships between them can be characterized using structured representations such as knowledge graphs \cite{ZhaoYXL22}. This operation enables querying formats beyond pairwise comparisons, allowing for feature-level preference elicitation \cite{AustinKTS24}. Similar strategies have been adopted in the context of LLMs, where human explanations and rationales offer valuable guidance during fine-tuning \cite{GhazimatinPRW21}. In language processing, natural language inference and LLMs have also been employed for relation augmentation and annotation \cite{DingQZLLCX0LJ24}. Graph-based representations, as supported by recent work and our findings, have the potential to offer a strong foundation for both structured feedback and preference optimization in IPE \cite{PanLWCWW24}.

\section{Details of Proposed Algorithms}
\label{sec:appendix_algorithm}
We integrate the augmented confidence bounds into two canonical DB algorithms: RUCB and DTS. They belong to different types of DB strategies, the RUCB is deterministic, while the DTS is stochastic. Different types of strategies apply to different IPE scenarios, for instance, DTS is observed more efficient for response optimization in LLM exploration \cite{DwaracherlaAHR24}.

The full pseudo-code of \texttt{IPEA-RUCB} is provided in Algorithm~\ref{alg:rucb-ipeahf}. Its core selection strategy follows the original RUCB algorithm, with additional components for dependency extraction and feedback augmentation. The overall structure of \texttt{IPEA-DTS} is analogous to that of \texttt{IPEA-RUCB}, differing primarily in its use of Thompson sampling with majority voting for arm selection.  For completeness, we refer readers to the original DTS paper \cite{WuL16} and our implementation; we omit the full pseudo-code of \texttt{IPEA-DTS} here for brevity.

\begin{algorithm}[htb]
    \caption{Pseudo-code of \texttt{IPEA-RUCB}}
   \label{alg:rucb-ipeahf}
    \begin{algorithmic}[1]
       \STATE {\bfseries Input:} Candidate number $K$, context set $\mathcal{X}$, $\alpha>0$, similarity threshold $\tau$, IPE budget $T$ 
       \STATE {\bfseries Init:}  Winning matrix $\mathbf{B} = [0]_{K\times K}$, similarity-based undirected graph $\mathcal{G}(\mathcal{X}, K)$ with edges determined by $\tau$, dependency dictionary $ W=[0]$, candidate set $\mathcal{B}=\emptyset$, $t=0$
       \REPEAT
       \STATE \textit{/ *** Implementation of AugConfidanceBound *** /}
       \FOR{$i,j\in \mathcal{A}$, $i \ne j$}
            \STATE Get tuples of related observations $\mathcal{T}_{n^r_{i,j}} \leftarrow (W,\mathcal{G}).\mathrm{get}(i,j)$, $n^r_{i,j} \leftarrow 0$, $n^r_{w} \leftarrow 0$, $X_{i,j} \leftarrow 0$
            \FOR{$(a_m,a_n, w_{i,j}^{m,n})\in \mathcal{T}_{n^r_{i,j}}$}
            \STATE $n^r_{i,j} \gets n^r_{i,j} + b_{m,n} + b_{n,m}$, $n_{w}^r \gets n_{w}^r + w_{i,j}^{m,n} * (b_{m,n} + b_{n,m})$, $X_{i,j} \gets X_{i,j} + b_{m,n}$
            \ENDFOR
            \STATE $\hat p_{i,j} = \frac{b_{i,j} + X_i}{b_{i,j} + b_{j,i} + n_w^r}$, $\hat u_{i,j} = \hat p_{i,j} + \sqrt{\frac{\alpha (b_{i,j} + b_{j,i} + n_{i,j}^r) \log t}{(b_{i,j} + b_{j,i} + n_w^r)^2}}$ \COMMENT{equation~\eqref{eqn:augmented-ucb}, $\frac{x}{0} := 1$, $\forall x$}
       \ENDFOR
       \STATE \textit{/ *** Implementation of DuelingBanditAlgo *** /} \COMMENT{RUCB algorithm in \cite{ZoghiWMR14}}
       \STATE $\mathcal{C} \gets \{a_c \vert \forall j: \hat u_{c,j} \ge 0.5\}$, \textbf{if} $\mathcal{C} = \emptyset$: \textbf{then} $\mathcal{C} \gets \{a_c \vert c = \mathrm{Random}(K)\}$, $\mathcal{B} \gets \mathcal{B} \cap \mathcal{C}$
       \STATE Sample $a_{c}$ from $\mathcal{C}$ according to distribution: \textbf{if} $a_c \in \mathcal {B}$: $p(a_c) = 0.5 $, \textbf{else} $p(a_c) = \frac{1}{2^{\vert \mathcal{B} \vert} \vert \mathcal{C} \setminus \mathcal{B} \vert} $
       \STATE Obtain $a_{d}$ where $d \gets \arg\max_j \hat u_{j,c}$, ties broken randomly with $d\ne c$
       \STATE Query user preference on $(a_c, a_d)$, with $X \sim \mathrm{Bernoulli}(p_{c,d})$
       \STATE \textbf{if} $X = 1$: $b_{c,d} \gets b_{c,d} + 1$ \textbf{else} $b_{d, c} \gets b_{d, c} + 1$
       \STATE \textit{/ *** Implementation of DependencyExtract \& FeedbackAug *** /}  
       \STATE Obtain related pairs $\{(a_m, a_n) \}$ from similarity graph $\mathcal{G}$ through connectivity
       \STATE Get additional annotations $w_{c,d}^{m,n}$ from user / LLMs
       \STATE Update $(W,\mathcal G).\mathrm{append}((c, d): (m, n, w_{c,d}^{m,n}))$ \COMMENT{add related observations with dependency} \\
       \STATE Get richer forms of feedback $F(a_c, a_d)$ and update $\mathcal G$ \COMMENT{this step is optional}\\
       \STATE $t \gets t + 1$ \\
       \UNTIL{$t> T$}
    \end{algorithmic}
\end{algorithm}

\section{Experimental Setup and Implementation Details}
\label{sec:appendix_exp}

\subsection{Benchmark Problems}

\paragraph{Sushi Dataset Details}
The Sushi dataset\footnote{\url{https://www.kamishima.net/sushi/}} collects full rankings over $10$ popular types of sushi from Japanese customers. It contains $5{,}000$ customer orders, along with item features and user demographic information. To construct the preference matrix, we traverse all orders and, for each pair of sushi items, count the number of times the higher-ranked item wins over the lower-ranked item. The resulting counts are then normalized by the total number of comparisons for each sushi pair. To incorporate contextual information, we leverage the item features listed in Table~\ref{tab:sushi_features}.

\begin{table}[htbp]
\centering
\caption{Item features for the Sushi dataset.}
\label{tab:sushi_features}
\begin{tabular}{lll}
\hline
\textbf{Feature} & \textbf{Type} & \textbf{Value Space} \\
\hline
Item ID         & Categorical     &  1-10 \\
Name            & Categorical     &  Roman alphabets of sushi type \\ 
Style           & Binary          &  0 (maki), 1 (otherwise) \\ 
Major Group     & Binary          &  0 (seafood), 1 (otherwise) \\ 
Minor Group     & Categorical     &  0 (blue skinned fish)-11 (vegetables) \\ 
Heaviness Taste & Numerical       & [0-4], 0: heavy \\
Eating frequency& Numerical       & [0-3], 3: most frequently \\
Price           & Numerical       & - \\
Selling Frequency & Numerical     & [0-1], 1: most frequently \\
\hline
\end{tabular}
\end{table}

\begin{figure*}[htb]
    \centering
    \includegraphics[width=0.9\linewidth]{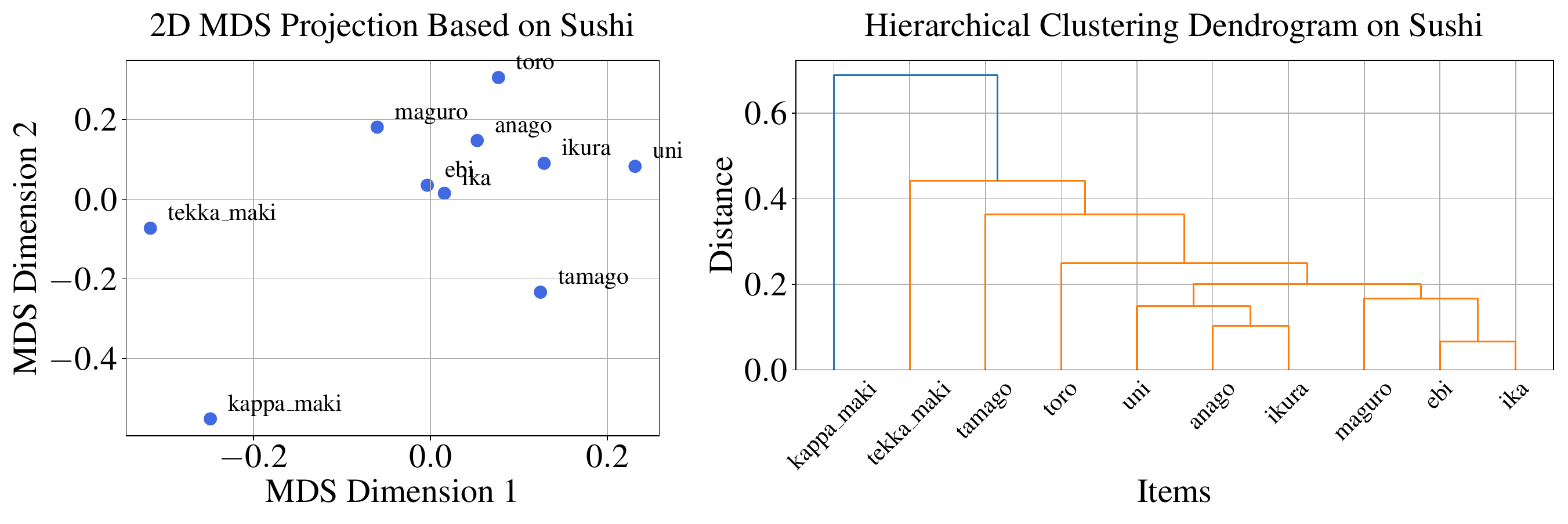}
    \caption{Hierarchical clustering dendrogram of Sushi items using Gower distance. Vertical distances indicate dissimilarity; branches closer together denote similar items.}
    \label{fig:sushi-dendrogram-gower}
\end{figure*}

Among all sushi types, toro (fatty tuna) is the most preferred item, identified as the Condorcet winner in the preference matrix, with the feature vector: [8, toro, 1, 0, 1, 0.55, 2.06, 4.49, 0.80]. For our analysis, we discard the ID and Name features and normalize numerical features (Heaviness Taste, Eating frequency, Price, Selling Frequency) consistently across all algorithms. Figure~\ref{fig:sushi-dendrogram-gower} presents both the 2-dimensional multidimensional scaling (MDS) projection and the hierarchical clustering dendrogram of sushi items based on their features. The figure suggests that sushi items do not clearly separate into distinct subsets, implying that partition-based methods might yield suboptimal performance in this scenario.

\paragraph{Car Preference Dataset Details} The Car Preference dataset \footnote{\url{https://ehsanabb.github.io/CarPreferences.html}} collects full ranking over $10$ types of cars from $60$ American users in Amazon Mechanical Turk. We conduct the same operation as the Sushi experiment to obtain the preference matrix. The item features are listed in Table~\ref{tab:car_features}.

\begin{table}[htbp]
\centering
\caption{Item features for the Car Preference dataset.}
\label{tab:car_features}
\begin{tabular}{lll}
\hline
\textbf{Feature} & \textbf{Type} & \textbf{Value Space} \\
\hline
Item ID         & Categorical     &  1-10 \\
Body Type       & Binary          &  1 (Sedan), 2 (SUV) \\ 
Transmission    & Binary          &  1 (Manual), 2 (Automatic) \\ 
Engine Capacity & Categorical     &  2.5L, 3.5L, 4.5L, 5.5L, 6.2L \\ 
Fuel Consumed   & Binary          &  1 (Hybrid), 2 (Non-Hybrid) \\ 
\hline
\end{tabular}
\end{table}

\begin{figure*}[htb]
    \centering
    \includegraphics[width=0.9\linewidth]{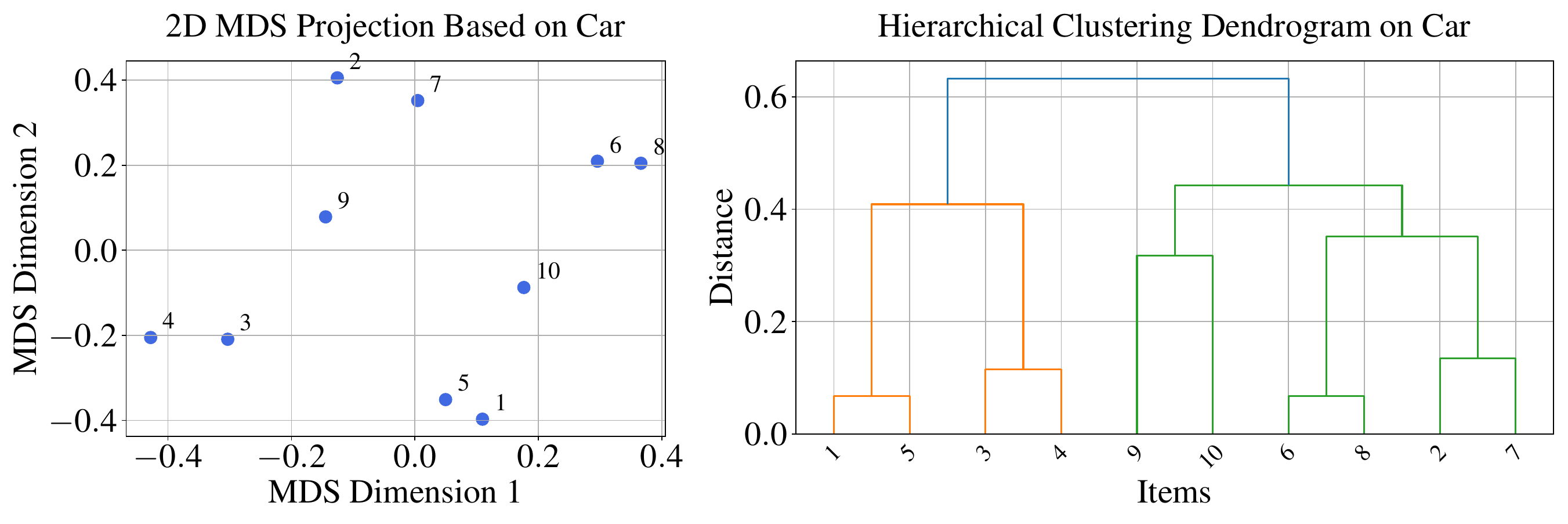}
    \caption{Hierarchical clustering dendrogram of Car items using Gower distance.}
    \label{fig:car-dendrogram-gower}
\end{figure*}

Among all car types, the sixth car is the most preferred item, identified as the Condorcet winner in the preference matrix, with the feature vector: [2, 2, 3.5L, 1]. 
Figure~\ref{fig:car-dendrogram-gower} presents both the 2-dimensional multidimensional scaling (MDS) projection and the hierarchical clustering dendrogram of sushi items based on their features. The figure suggests that car items may be clearly separated into distinct subsets (4 subsets and two single items 9 and 10).

\begin{figure*}[htbp]
  \centering
  \begin{subfigure}[b]{0.45\textwidth}
    \centering
    \includegraphics[width=\linewidth]{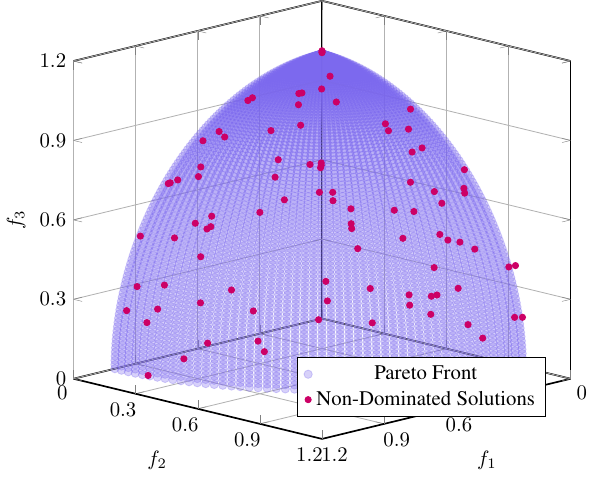}
    \caption{Pareto front and samples of DTLZ2.}
  \end{subfigure}
  \hfill
  \begin{subfigure}[b]{0.45\textwidth}
    \centering
    \includegraphics[width=\linewidth]{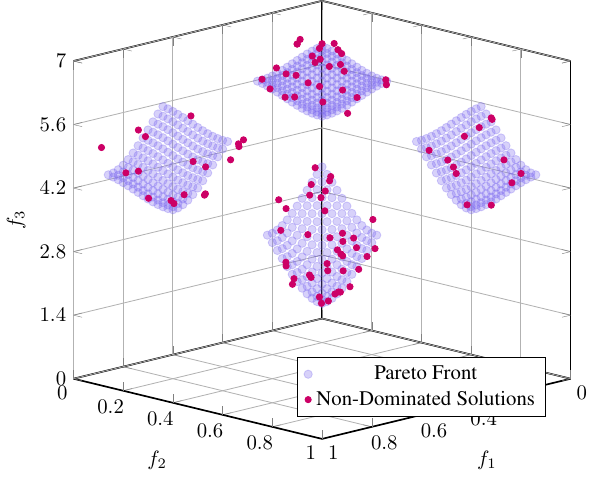}
    \caption{Pareto front and samples of DTLZ7.}
    \label{fig:subB}
  \end{subfigure}
  \caption{DTLZ Pareto front with $100$ samples in $3$-dimensional objective space.}
  \label{fig:dtlz}
\end{figure*}

\paragraph{Synthetic DTLZ}
The synthetic DTLZ problems are commonly used for benchmarking in research of multi-objective optimization \cite{HuangWL24}. We select two DTLZ problems (DTLZ2 and DTLZ7) with different Pareto fronts as visualized in Figure~\ref{fig:dtlz}. Note that DTLZ 1-4 share the same characteristics in Pareto fronts, while DTLZ 5 and DTLZ 6 is much simpler than other problems in DTLZ family in terms of preference optimization. 

We use the \textit{a posteriori} method of multi-objective optimization to focus on the IPE task. We sample $100$ non-dominant solutions in objective space using NSGA-II, one of the most prestigious evolutionary-based algorithm. Following the settings in \cite{HuangWL24}, we randomly select one sample as the best winner, denoted as $\vec x_*$. Then, we construct the preference matrix by
\begin{equation}
    p_{i,j} = \mathrm{sigmoid} \left(\frac{u_i - u_j}{\tau}\right), \quad u_i = \mathcal{N}(\vec x_i \vert \vec x_*, \sigma\mathbf{I}), \quad i,j\in \{1, 2,\dots, 100\}.
\end{equation}
Here, $\vec{x}_i$ denotes the decision variable of a sampled solution. We set the preference noise threshold to $\tau = 0.2$ to generate clearer pairwise preferences. The DTLZ problem involves a 10-dimensional decision variable and a 3-dimensional objective vector. We concatenate these two components to form a 13-D feature vector for each solution. This unified representation is used by the parametric DB algorithms and serves as the input for similarity computation within our framework.

\begin{figure*}[htb]
    \centering
    \includegraphics[width=\linewidth]{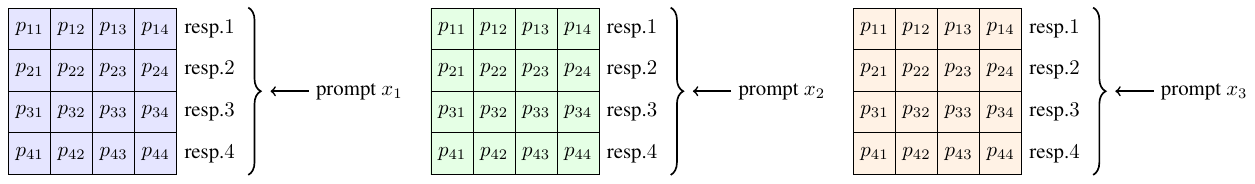}
    \caption{Illustration of preference matrices in response optimization. This represents a contextual dueling bandit (DB) setting, where each context corresponds to a different prompt, and is associated with a distinct pool of response candidates and its own preference matrix.}
    \label{fig:llm-matrices}
\end{figure*}

\paragraph{LLM Exploration}
We follow the official codebase \footnote{\url{https://github.com/xqlin98/APOHF}} provided by the authors of \cite{linDV24}. The prompt optimization task can be formulated as a contextual DB problem. Specifically, given a prompt $x_i$, instead of generating two responses randomly for preference optimization, one can generate a large pool of candidate responses and actively select the one mostly aligned with human preferences. This process is also referred to as active exploration in the literature \cite{DwaracherlaAHR24}. As depicted in Figure~\ref{fig:llm-matrices}, this task fits naturally within the contextual DB setting \cite{LattimoreS20}, where each prompt (context) is associated with a preference matrix, and the objective is to identify the optimal arm given the current context.

We randomly sample $5$ prompts from the Anthropic Helpfulness and Harmlessness datasets as different contexts, and generate $20$ responses for each prompt using GPT-3.5-turbo. To mimic human preferences, we leverage a fine-tuned reward model based on GPT-2 trained on the same datasets \footnote{\url{https://huggingface.co/Ray2333/gpt2-large-helpful-reward_model}} as the reward function $f$. Pairwise winning probabilities are computed using  the BT model. Note that for a given prompt, only responses from the same pool are valid for comparisons, while inter-prompt comparisons are meaningless for IPE. Consequently, standard DB algorithms without explicit mechanisms for dealing with inter-prompt contextual information \cite{linDV24,VermaDLJL25} must maintain $5$ separate winning matrices or parametric models and select the candidates based on current prompt/context. Our algorithms face the same difficulty, but are able to go further: they leverage augmented feedback not only from responses within the same prompt/context, but also from other prompts, provided strong contextual similarity and dependency can be established. Since these augmented observations originate from different candidate pools, we treat them as a distinct form of human feedback.

\subsection{Algorithmic Details and Configurations}
\paragraph{Peer Algorithms}
We use the official implementation of \texttt{VACDB}\footnote{\url{https://github.com/uclaml/VACDB}}, which also include the implementation of
\texttt{MaxInP} and \texttt{COLSTIM}. We keep the default setting of the three algorithms. 
We implement RUCB and DTS in Python according to descriptions in literature.

For mixed-valued feature spaces (e.g., Sushi and Car Preference datasets), we compute pairwise similarity using the default Gower distance. For datasets with purely numerical features (e.g., DTLZ and LLM Exploration), we apply Euclidean distance after normalization with \texttt{MinMaxScaler} from \texttt{sklearn}. The similarity threshold for feedback augmentation is set to $0.85$ across all experiments. For LLM-based annotation, we set the temperature as $0.9$ to encourage variability. We use a fixed confidence parameter of $\alpha = 0.1$ for \texttt{RUCB}, \texttt{DTS}, and our proposed algorithms. The sensitivity of our method to different $\alpha$ values is further analyzed in Section~\ref{sec:app_exp_para_alpha}.

\paragraph{Experimental Configurations}
we run all experiments on a MacBook Pro equipped with an Apple M3 Pro chip and 36 GB of RAM. For each benchmark, every algorithm is executed with 10 different random seeds. The regret trajectories reported in all figures represent the mean regret, with shaded regions denoting $\pm0.5$ standard deviation. All implementations are written in Python, and major dependencies (e.g., \texttt{numpy}, \texttt{scikit-learn}, \texttt{gower}, and \texttt{networkx}) are publicly available. To speed up the experiments, we pre-process all benchmarks into a unified data format, including the preference matrix, similarity graph, and dependency dictionary.

\subsection{Prompt Design for Dependency Annotation}
We use the following prompts for LLM annotator (GPT-3.5-Turbo) to reason conditional dependency.

\begin{tcolorbox}[colback=gray!5, colframe=gray!30, title=Prompt for Sushi Recommendation]
You are assisting in eliciting user preferences in the context of sushi recommendation. 
\\

Here are four sushi types with their descriptions:

- Item A: ebi, style not maki, major group seafood, minor group blue-skinned fish, ...

- Item B: anago, style not maki, major group seafood, minor group red meat fish, ...

- Item C: ebi, style not maki, major group seafood, minor group blue-skinned fish, ...

- Item D: ikura, style not maki, major group seafood, minor group squid or octopus, ...
\\

Assume the following:
The user has shown a strong preference for Item A over Item B.

You should consider both the features and your experience. 
\\

Please reason briefly about how this affects the likelihood of preferring Item C over Item D, and then on a separate line, provide only the numeric score from 0 to 1.

\end{tcolorbox}

Accordingly, one piece of the response from GPT is:

\begin{tcolorbox}[colback=blue!5, colframe=blue!40, title=LLM Response to Sushi Preference Dependency (GPT 3.5 Turbo)]
Given that the user has a strong preference for Item A over Item B, 
it is likely that they would also prefer Item C over Item D 
since both items share similar characteristics with Item A in terms of 
the type of seafood (ebi) and eating frequency. Additionally, both 
Item A and Item C have a higher selling frequency compared to Item B and D,
indicating they are more popular choices among consumers.

Numeric score: 0.8
\end{tcolorbox}

In Car recommendation, we use the following prompt structure:

\begin{tcolorbox}[colback=gray!5, colframe=gray!30, title=Prompt for Car Recommendation]
You are assisting in eliciting user preferences in the context of car recommendation. 
\\

Here are four car types with their descriptions:

- Item A: body type SUV, transmission Manual, engine capacity 2.5, ...

- Item B: body type Sedan, transmission Automatic, engine capacity 3.5, ...

- Item C: body type SUV, transmission Manual, engine capacity 3.5, ...

- Item D: body type Sedan, transmission Automatic, engine capacity 3.5, ...
\\

Assume the following:
The user has shown a strong preference for Item A over Item B.
\\

You should consider both the features and your experience. Please reason briefly about how this affects the likelihood of preferring Item C over Item D, and then on a separate line, provide the value of numeric score in [0, 1].

\end{tcolorbox}

Accordingly, one piece of the response from GPT is:

\begin{tcolorbox}[colback=blue!5, colframe=blue!40, title=LLM Response to Car Preference Dependency (GPT 3.5 Turbo)]
Given that the user has a strong preference for Item A over Item B, it is likely that they value the body type (SUV) and transmission type (Manual) over engine capacity and fuel consumption. Therefore, it is possible that they may also prefer Item C over Item D, as it shares the same body type and transmission type as Item A. 

Numeric score: 0.75
\end{tcolorbox}

\section{Experimental Results}

\subsection{Additional Results}
\begin{figure*}[htb]
    \centering
    \includegraphics[width=1\linewidth]{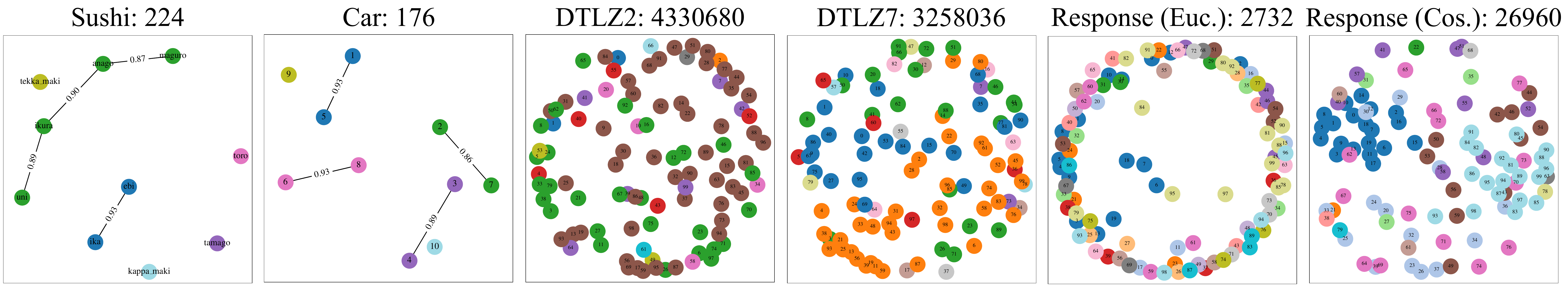}
    \caption{Similarity graph and total size of dependency dictionaries for each benchmark, under a similarity threshold of $0.85$. In LLM response, we use both Euclidean and Cosine distances.}
    \label{fig:graph}
\end{figure*}

\paragraph{Similarity Graph}
The similarity graph guides the algorithms to focus on the most informative candidate pairs, represented by the edges, which are likely to yield meaningful dependency information. The constructed graphs for each benchmark are visualized in Figure~\ref{fig:graph}. The graph structures for Sushi and Car are consistent with the similarity analyses presented in Figures~\ref{fig:sushi-dendrogram-gower} and \ref{fig:car-dendrogram-gower}. Notably, the size of the resulting dependency dictionaries varies across benchmarks. This is particularly evident in the response optimization task, where only comparisons among responses under the same prompt are retained, imposing an additional structural constraint that is absent in the DTLZ problems.

\begin{figure*}[htb]
    \centering
    \includegraphics[width=\linewidth]{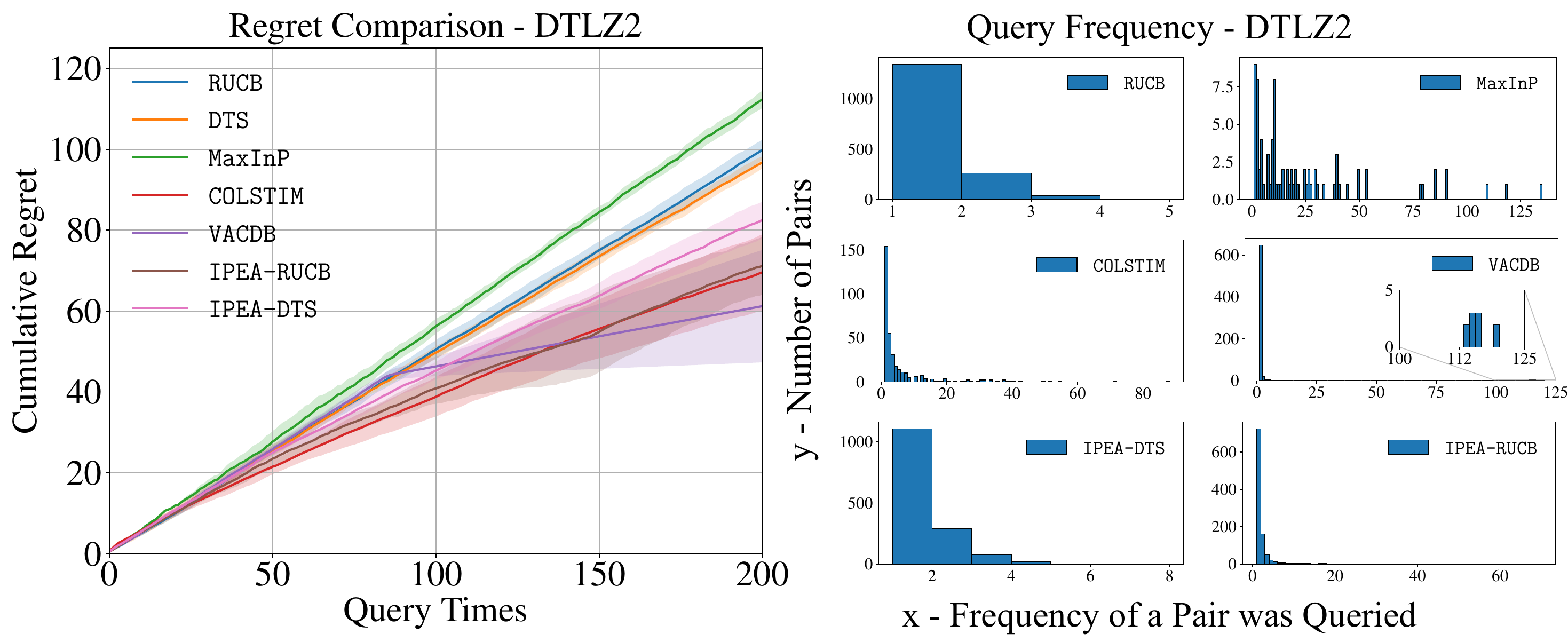}
    \caption{Regret and query frequency analysis on DTLZ2.}
    \label{fig:mo2_result}
\end{figure*}

\paragraph{Multi-Objective Optimization} 
The results on DTLZ2 problem are given in Figure~\ref{fig:mo2_result}. We also present numerical scores of each algorithm in Table~\ref{tab:query-stats-all}, demonstrating a better trade-off of our algorithms between regret minimization and query diversity.

\begin{table*}[ht]
\small
\centering
\caption{Query Statistics Across Algorithms in DTLZ Benchmarks}
\begin{tabular}{@{}ll *{7}{r}@{}}
\toprule
\textbf{Exp.} & \textbf{Metric} & \texttt{RUCB} & \texttt{DTS} & \texttt{MaxInP} & \texttt{COLSTIM} & \texttt{VACDB} & \texttt{IPEA-RUCB} & \texttt{IPEA-DTS} \\
\midrule
\multirow{4}{*}{DTLZ2} 
& Total Queries         & 2000 & 2000 & 2000 & 2000 & 2000 & 2000 & 2000 \\
& Unique Pairs          & 1651  & 1658 & 83  & 351 & 687 & 1020 & 1496 \\
& Entropy (bits)        & 10.59  & 10.60 & 5.53 & 7.20 & 6.81 & 9.21 & 92.2\% \\
& Normalized Entropy    & 99.1\%  & 99.2\% & 86.8\% & 85.1\% & 72.3\% & 92.2\% & 98.7\% \\
\midrule
\multirow{4}{*}{DTLZ7} 
& Total Queries         & 2000  & 2000 & 2000 & 2000 & 2000 & 2000 & 2000 \\
& Unique Pairs          & 1650  & 1668 & 89 & 171 & 484 & 875 & 1568 \\
& Entropy (bits)        & 10.59 & 10.62& 5.71 & 5.95 & 4.10 & 9.27 & 10.49 \\
& Normalized Entropy    & 99.1\%& 99.2\%& 88.2\% & 80.2\% & 46.0\% & 94.8\% & 98.9\% \\
\bottomrule
\end{tabular}
\label{tab:query-stats-all}
\end{table*}

\subsection{Parameter Sensitivity}
\label{sec:app_exp_para_alpha}
\begin{figure*}[htb]
    \centering
    \includegraphics[width=\linewidth]{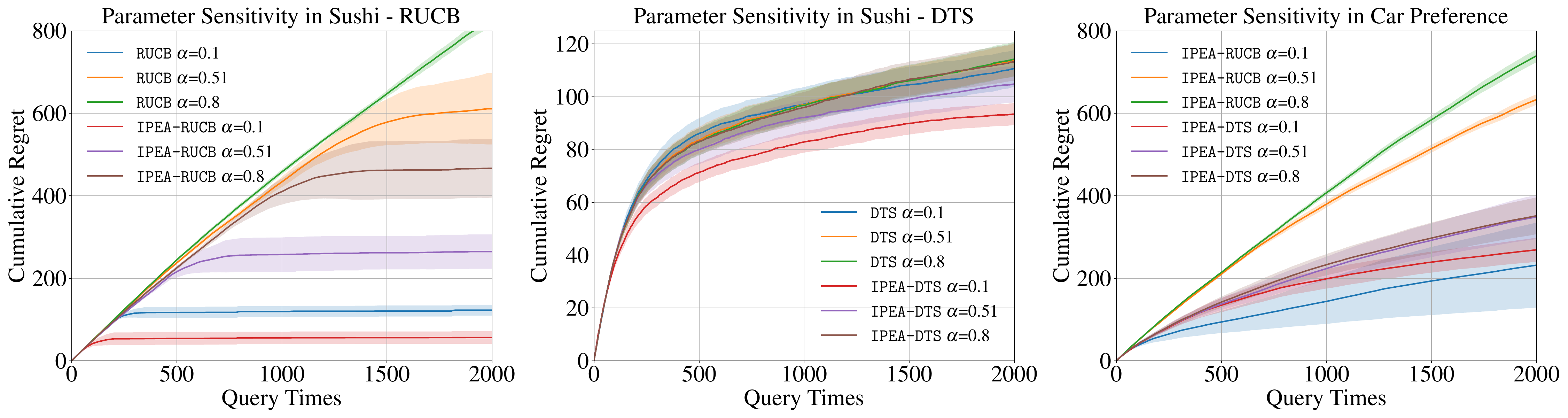}
    \caption{Regret trajectories for Sushi and Car Preference experiments under different values of $\alpha$.}
    \label{fig:regret_alpha}
\end{figure*}

We study the sensitivity of the exploration parameter $\alpha$ in both \texttt{IPEA-RUCB} and \texttt{IPEA-DTS}. This parameter controls the degree of exploration: higher values of $\alpha$ encourage more exploratory behavior. To assess its effect on convergence, the we choose the recommendation tasks using three representative values of $\alpha$ from the set $[0.1, 0.51, 0.8]$. The comparative results are presented in Figure~\ref{fig:regret_alpha}.

Our findings suggest that the deterministic strategies, such as \texttt{RUCB} and \texttt{IPEA-RUCB}, are more sensitive to changes in $\alpha$. This sensitivity reflects a potential weakness of deterministic strategies as they may be more susceptible to suboptimal parameter settings, in contrast to stochastic strategies (e.g., \texttt{DTS} and \texttt{IPEA-DTS}), as has been noted in \cite{WuL16, HuangWL24, DwaracherlaAHR24}. However, \texttt{IPEA-RUCB} consistently outperforms the behaviors of original \texttt{RUCB}, demonstrating the effectiveness of feedback augmentation. 

In contrast, both \texttt{DTS} and \texttt{IPEA-DTS} exhibit minimal sensitivity to $\alpha$, with stable performance across the tested range. Based on our experiments, a fixed setting of $\alpha = 0.1$ proves effective for \texttt{IPEA-RUCB} in most cases. For \texttt{IPEA-DTS}, $\alpha = 0.1$ is also recommended, though other values yield comparably strong performance.

\newpage

\section{Theoretical Analysis}
\label{sec:appendix_theoretical}
In this section, we present formal proofs and detailed derivations for our main theoretical results. We begin by introducing a general form of the Chernoff–Hoeffding inequality as a foundational lemma. Building on this, we prove Theorem~\ref{thm:concentration}, which establishes the concentration property under augmented feedback. We then derive the calibration threshold that characterizes when augmented feedback contributes positively to concentration. Subsequently, we analyze the total sample complexity for each pair under the \texttt{IPEA-RUCB} algorithm, supporting the claims stated in Theorem~\ref{thm:sample_number}. Finally, we provide the proof of Theorem~\ref{thm:regret}, which characterizes the regret bound within \texttt{IPEA-RUCB}.

\begin{lemma}[Chernoff-Hoeffding inequality]
\label{lemma:appendix_CF_bound}
    Suppose that random variables $X_1,\dots,X_n$ are independent, $a_i \leq X_i \leq b_i$, and let $E[X_i]=\mu$. Then for any $\sigma>0$, the following inequality holds:
    \begin{equation}
        P\left(\left\vert \frac{1}{n} \sum_{i=1}^n X_i - \mu \right\vert \ge \sigma \right) \leq 2\exp \left( -\frac{2n^2 \sigma^2}{\sum_{i=1}^n(b_i - a_i)^2} \right).
        \label{eqn:lemma1-CH}
    \end{equation}
\end{lemma}

This inequality applies to independent random variables that are not necessarily identically distributed, a property that is particularly useful for our theoretical analysis with augmented human feedback. In the context of dueling bandits, where the random variables are Bernoulli-distributed (i.e., $a_i = 0$ and $b_i = 1$ for $i = 1, \dots, n$), the inequality can be formulated as follows:
\begin{equation}
        P\left( \frac{1}{n} \sum_{i=1}^n X_i \ge \mu + \sigma \right) \leq \exp \left( -{2n \sigma^2} \right), \quad 
        P\left( \frac{1}{n} \sum_{i=1}^n X_i \leq \mu - \sigma \right) \leq \exp \left( -{2n \sigma^2} \right).
\end{equation}

\subsection{Concentration Property: Proof of Theorem~\ref{thm:concentration}}
\label{sec:appendix_theoretical_concentration}

\begin{proof}
We follow the steps in \cite{ZoghiWMR14} by first introducing the notation for the \textit{good} event $\mathcal{E}_{i, j}(t)$, which denotes that at time $t$, the true preference probability satisfies $ p_{i, j } \in \left[ \hat l_{i,j}(t),~ \hat u_{i, j}(t) \right]$. Note that $\mathcal{E}_{i, i}(t)$ holds for all $t$ and that $\mathcal{E}_{i, j}(t) = \mathcal{E}_{j, i}(t)$ for all $i, j \in \mathcal{A}$. Therefore, we focus without loss of generality on the case where $i > j$. 
% Define $\tau^n_{i,j}$ as the $n$-th number of time when $a_i$ and $a_j$ are compared against each other or related observations are received from other comparisons. 
We define $\tau^n_{i,j}$ as the time at which the $n$-th total observation for the pair $(a_i, a_j)$ occurs. Here, the total observations include both direct comparisons between $a_i$ and $a_j$, and related observations inferred from comparisons involving other arms.
Another fact is that for $T^\prime \in [\tau^n_{i,j}, \tau^{n+1}_{i,j}]$, if $\mathcal{E}_{i, j}(T^\prime)$ holds, then $\mathcal{E}_{i, j}(t)$ also holds  for all $t \in [T^\prime, \tau^{n+1}_{i,j}]$. This is true because for $t \in [\tau^n_{i,j}, \tau^{n + 1}_{i,j}]$, the uncertainty terms in $\hat u_{i,j}$ and $\hat l_{i,j}$ increase monotonically. Consequently, for any $T^\prime$ and all subsequent times $t > T^\prime$, the probability of event $\mathcal{E}_{i, j}(t)$ occurring is expressed as follows:
\begin{equation}
     P\left(\forall t \ge T^\prime, i, j, ~ \mathcal{E}_{i, j}(t)\right)
    =  P\left(\forall i > j, ~\mathcal{E}_{i, j}(T^\prime) \land  \forall n ~s.t.~ \tau^{n}_{i,j}>T^\prime,~ \mathcal{E}_{i, j}(\tau^{n}_{i,j}) \right).
\end{equation}
Conversely, the \textit{bad} event $\mathcal{B}_{i, j}(t)$, defined as the event that $p_{i, j } \notin \left[ \hat l_{i,j}(t),~ \hat u_{i, j}(t) \right]$ at the $t$-th iteration, can be bounded by the following inequality, applying notions from equation~\eqref{eqn:augmented-ucb}:
\begin{equation}
\begin{split}
    & P\left(\exists t \ge T^\prime, i, j, ~\mathcal{B}_{i, j}(t)\right) 
    =  1 - P\left(\forall t \ge T^\prime, i, j,~ \mathcal{E}_{i, j}(t)\right) \\ 
    \leq & \sum_{i<j} \left[ P\left( \mathcal{B}_{i, j}(T^\prime) \right) + P\left( \exists n ~s.t.~\tau^{n}_{i,j} > T^\prime, ~\mathcal{B}_{i, j}(\tau^{n}_{i,j}) \right) \right] \\
     \leq & \sum_{i<j}\left[ 2\sum_{n=0}^{T^\prime} P \left( \left\vert p_{i,j} - \hat p_{i,j}(n) \right\vert > \frac{1}{\eta(n)}\sqrt{\frac{\alpha \ln T^\prime}{n}} \right)\right. \\
     & \quad \quad \left. + \sum_{n=T^\prime+1}^\infty P\left( \left\vert p_{i,j} - \hat p_{i,j}(n) \right\vert > \frac{1}{\eta(n)}\sqrt{\frac{\alpha \ln n}{n}} \right) \right] .
\end{split}
\label{eqn:proof1-badevent}
\end{equation}

The final step of the above inequalities holds as proven in \cite{ZoghiWMR14}. For conciseness, we let $\hat p_{i,j}({n})$ denote the estimate of $p_{i,j}$ given $n$ total observations. Similarly, we explicitly denote the dependency of $\eta$ on $n$ by writing it as $\eta(n)$.

On the other hand, given $n_{i,j}^d$ independent observations drawn from $\mathrm{Bernoulli}(p_{i,j})$ and $n_{i,j}^r$ additional independent observations drawn from separate distributions $\mathrm{Bernoulli}(w_{i,j}^k p_{i,j})$ for $k = 1, \dots, n_{i,j}^r$, the expectation of the combined observations can be expressed as follows:
\begin{equation}
    \mu = \frac{n_{i,j}^d}{n_{i,j}}p_{i,j} + \frac{1}{n_{i,j}} \sum_{k=1}^{n_{i,j}^r}  w^k_{i,j} p_{i,j} =  \frac{1}{n_{i,j}}\left(n_{i,j}^d + \sum_{k=1}^{n_{i,j}^r} w^k_{i,j}\right) p_{i,j} :=   \eta p_{i,j}.
\end{equation}
By applying the generalized Chernoff-Hoeffding inequality in equation~\eqref{eqn:lemma1-CH}, we have 
\begin{equation}
    P\left(\left\vert \frac{1}{n_{i,j}} \left(b_{i,j} + \sum_{k=1}^{n^r_{i,j}} X_{i,j}^k\right) - \eta p_{i,j} \right\vert \ge \sigma \right) \leq 2\exp \left( -{2n_{i,j} \sigma^2} \right),
\end{equation}
for any $\sigma>0$. Dividing both sides of the inequality inside the probability by $\eta$, we obtain
\begin{equation}
    P\left(\left\vert \hat p_{i,j} - p_{i,j} \right\vert \ge \sigma^\prime \right) \leq 2\exp \left( -{2\eta^2n_{i,j}  \sigma^{\prime 2}} \right),
    \label{eqn:proof1-chbound-trans}
\end{equation}
where $\sigma^\prime := \sigma / \eta$. Combining the results in inequalities~\eqref{eqn:proof1-badevent} and \eqref{eqn:proof1-chbound-trans} by letting $\sigma^\prime$ be $\frac{1}{\eta(n)} \sqrt{\frac{\alpha \ln T^\prime}{n}}$ or $\frac{1}{\eta(n)} \sqrt{\frac{\alpha \ln n}{n}}$ in each term of \eqref{eqn:proof1-badevent}, we further have
\begin{equation}
\begin{split}
    P\left(\exists t \ge T^\prime, i, j, ~\mathcal{B}_{i, j}(t)\right)
     &\leq \sum_{i<j}\left[ 4\sum_{n=0}^{T^{\prime}} e ^{-2\alpha\ln T^\prime} + 2\sum_{n=T^\prime+1}^\infty e^{-2\alpha\ln n} \right] \\
     &= \frac{K(K-1)}{2}\left[ \sum_{n=1}^{T^{\prime}} \frac{4}{T^{\prime~2\alpha}} + \sum_{n = T^\prime +1}^\infty \frac{2}{n^{2\alpha}}\right].
\end{split}
\end{equation}
This result align with that in \cite{ZoghiWMR14}. By treating $n$ as a continuous variable, we can further bound it by
\begin{equation}
    P\left(\exists t \ge T^\prime, i, j, ~\mathcal{B}_{i, j}(t)\right)
     \leq \frac{(4\alpha - 1)K^2}{(2\alpha - 1)T^{\prime ~ 2\alpha - 1}}, \text{ with } T^{\prime~2\alpha - 1} > 0.
\end{equation}
By setting $T^\prime = C(\delta)$, inequality~\eqref{eqn:thm1-concentration} holds, which completes the proof.
\end{proof}

\subsection{Calibration Threshold}
\label{sec:appendix_theoretical_threshold}
Here we present the derivations of the ratio and the calibration threshold in equations \eqref{eqn:ratio} and \eqref{eqn:cali_threshold}.

\paragraph{Ratio of Confidence Intervals}
Considering $n_{i,j}^r(t)=1$, at the $t$-th round, the uncertainty ratio of confidence bounds with and without a relative observation is
\begin{equation}
   \frac{\hat u_{i,j} - \hat l_{i,j}}{\tilde u_{i, j} - \tilde l_{i, j} }  =\frac{1}{\eta} \sqrt{\frac{\alpha \ln t}{n^d_{i,j}(t)+ 1}}\cdot \sqrt{\frac{n^d_{i,j}(t)}{\alpha \ln t}} = \frac{1}{{\eta}} \sqrt{\frac{n_{i,j}^d(t)}{n_{i,j}^d(t)+1}} = \sqrt{\frac{n_{i,j}^d(t)(n_{i,j}^d(t)+1)}{(n_{i,j}^d(t)+w_{i,j}^r)^2}}.
\end{equation}
This ratio is inversely proportional to $w_{i,j}^r$. For $w_{i,j}^r = 0$ and $w_{i,j}^r = 1$, we have
\begin{equation}
\begin{split}
   \sup \frac{\hat u_{i,j} - \hat l_{i,j}}{\tilde u_{i, j} - \tilde l_{i, j} }  = \sqrt{\frac{n_{i,j}^d(t)+1}{n_{i,j}^d(t)}} = \sqrt{1 + \frac{1}{n_{i,j}^d(t)}}, \\
   \inf \frac{\hat u_{i,j} - \hat l_{i,j}}{\tilde u_{i, j} - \tilde l_{i, j} }  = \sqrt{\frac{n_{i,j}^d(t)}{n_{i,j}^d(t)+1}} = \sqrt{1 - \frac{1}{n_{i,j}^d(t)+1}},
\end{split}
\end{equation}
which indicates the interval in equation~\eqref{eqn:ratio}.

\paragraph{Calibration Threshold}

Given $n_{i,j}^d$ direct observations and $n_{i,j}^r$ related observations, we consider the case where a new related observations can be added. The ratio of interval with (denoted by $\hat u^{1}_{i,j} - \hat l^{1}_{i,j}$ and $\eta^{1}$) and without ( denoted by $\hat u^{0}_{i,j} - \hat l^{0}_{i,j}$ and $\eta^0$) the added observation becomes:

\begin{equation}
   \frac{\hat u^{1}_{i,j} - \hat l^{1}_{i,j}}{\hat u^{0}_{i,j} - \hat l^{0}_{i,j}}  =\frac{\eta^0}{\eta^1} \sqrt{\frac{\alpha \ln t}{n_{i,j}(t)+ 1}}\cdot \sqrt{\frac{n_{i,j}(t)}{\alpha \ln t}} = \frac{\eta^0}{\eta^1} \sqrt{\frac{n_{i,j}(t)}{n_{i,j}(t)+1}} = \frac{\theta}{\theta + w_{i,j}^r}\sqrt{\frac{n_{i,j}(t) + 1}{n_{i,j}(t)}} ,
\end{equation}
where $\theta = \eta_0 n_{i,j}(t)$. To ensure a positive contribution, the ratio should less than $1$, thus we have
\begin{equation}
    \frac{\theta}{\theta + w_{i,j}^r}\sqrt{\frac{n_{i,j}(t) + 1}{n_{i,j}(t)}}  \leq 1~ \Rightarrow ~ w_{i,j}^r \ge \theta \left( \sqrt{\frac{n_{i,j}(t) + 1}{n_{i,j}(t)}} - 1 \right) = \theta \left( \sqrt{1 + \frac{ 1}{n_{i,j}(t)}} - 1 \right),
\end{equation}
which completes the derivation.

\subsection{Sample Complexity: Proof of Theorem~\ref{thm:sample_number}}
\label{sec:appendix_theoretical_compare_number}

\begin{proof}
    For the case $i = j$, an arm cannot be played against itself with probability $1-\delta$ when $t > C(\delta)$ according to Theorem~\ref{thm:concentration}. It is because when the first arm is selected as the first candidate in \texttt{IPEA-RUCB}, the second arm cannot be the same arm given the fact that $\hat u_{i,j} = 0.5 < p_{*, i} \leq \hat u_{*, i}$, $\forall t > C(\delta)$. This ensures that self-comparisons are avoided with high probability once the concentration condition holds.
    Consider the case where $i \ne j$. Let $s$ be the last time a direct or related observation was added to the total observations of pair $(a_i, a_j)$, we have 
    \begin{equation}
    \label{min_delta}
        \hat u_{i,j}(s)-\hat l_{i,j}(s)= \frac{2}{\eta} \sqrt{\frac{\alpha \ln s}{n_{i,j}}}\leq \sqrt{\frac{4\alpha\ln t}{\min_r{w_{i,j}^{r~2}} \cdot n_{i,j}}}.
    \end{equation}
    The last inequality holds based on the fact that $w_{i,j}^r \in [0,1]$ for all $r$, and 
    \begin{equation*}
        \eta = \frac{n^d_{i,j} + \sum_{k=1}^{n^r_{i,j}} w^k_{i,j}}{n_{i,j}} \ge \frac{\min_r w^r_{i,j} \cdot n^d_{i,j} + \sum_{k=1}^{n^r_{i,j}} \min_r w^r_{i,j}}{n_{i,j}} = \min_r w^r_{i,j} \frac{n^d_{i,j} + n_{i,j}^r}{n_{i,j}} = \min_r w^r_{i,j}.
    \end{equation*}
    Thus, suppose $n_{i,j} \ge D_{i,j}^w\ln t$ and $t>C(\delta)$ with $D_{i,j}^w$ defined in Theorem~\ref{thm:sample_number}, we must have
    \begin{equation}
        \hat u_{i,j}(s)-\hat l_{i,j}(s) \leq \sqrt{\frac{4\alpha\ln t}{\min_r{w_{i,j}^{r~2}} \cdot n_{i,j}}} \leq \min \{\Delta_i^2,\Delta_j^2\}.
        \label{eqn:thmproof-u-l}
    \end{equation}
    % Different from \cite{ZoghiWMR14}, we need to consider the selection of both direct observation and related observation. 
    For a direct comparison of pair $(a_i, a_j)$, one of the following two scenarios must have happened:
    \begin{enumerate}[label=\textbf{Case \arabic*}]
        \item ($a_c= a_i$ and $a_d=a_j$) o include $a_i$ in the candidate set $\mathcal{C}$, it must hold that $\hat u_{i,j}(s) \geq \frac{1}{2}$. Meanwhile, we have $\hat l_{i,j}(s) = 1 - \hat u_{j,i}(s) \leq 1 - p_{*,i} $. This follows from two conditions: $i$) $p_{*,i} \leq \hat u_{*,i}(t)$ when $t > C(\delta)$, and $ii$) if \texttt{IPEA-RUCB} chooses $a_j$ as the most likely candidate to defeat $a_i$, we must have $\hat u_{*,i}(t) \leq \hat u_{j,i}(t)$. Combining these two inequalities yields: 
        \begin{equation}\label{delta_i}
            \hat u_{i,j}(s) - \hat l_{i,j}(s) \geq \frac{1}{2} - (1 - p_{*,i}) =p_{*,i} - \frac{1}{2}  = \Delta_i.
        \end{equation}
        \item ($a_c= a_j$ and $a_d=a_i$) By symmetry, swapping $i$ and $j$ in the above argument yields the corresponding result:
        \begin{equation}\label{delta_j}
            \hat u_{j,i}(s) - \hat l_{j,i}(s) \geq \frac{1}{2} - (1 - p_{*,j}) = \Delta_j.
        \end{equation}
    \end{enumerate}
    These two scenarios yield $\hat u_{i,j} - \hat l_{i,j} \ge \max\{\Delta_i^2, \Delta_j^2\}$, which contradicts inequality~\eqref{eqn:thmproof-u-l}. Therefore, with probability at least $1 - \delta$, it is not possible for $n_{i,j}$ to exceed both $C(\delta)$ and $D_{i,j}^w \ln t$.
    %%%%%%%%%%%%%% VERY IMPORTANT %%%%%%%%%%%%%%%%%%
    % For a relative observation, without loss of generality, a pair $(a_m, a_n)$ is selected to compare. Following the above derivation, it implies that $\hat u_{m,n} - \hat l_{m,n} \ge \max\{\Delta_m^2, \Delta_n^2\}$. We consider the worst case, that $n_{m,n} = n_{m,n}^d$, meaning pair $(a_m, a_n)$ does not use any related observations, and has the least dependency in $(a_i, a_j)$, i.e., $\min_r w_{i,j}^r$. Since we have already assumed $n_{i,j} \ge D^w_{i,j}$, by definition we have $n_{i,j} > n_{i,j}^d + n^d_{m,n}$.
\end{proof}

\subsection{Regret Analysis: Proof of Theorem~\ref{thm:regret}}
\label{thm:appendix_theoretical_regret}

\begin{proof}
     While the total sample complexity accounts for both direct and related observations, the cumulative regret is determined solely by direct observations. By definition, it is given by:
     \begin{equation}
        R(T) =  \sum_{i,j \in \mathcal{A},~i< j} n^d_{i,j}(T) \cdot \frac{\Delta_{i} + \Delta_{j}}{2}
         \leq \sum_{i,j \in \mathcal{A},~i< j} n^d_{i,j}(T).
    \end{equation}

    We decompose the cumulative regret $R(T)$ into multiple pair-level components, including inter-cluster pairs and intra-cluster comparisons among $C$ possibly overlapped clusters. An illustrative example is provided in Figure~\ref{fig:regret_illustration} to facilitate understanding. For inter-cluster pairs, any two pairs belong to two distinct clusters share the same number of observations. For instance, the pairs $(a_1, a_4)$ and  $(a_2, a_4)$ share the same observation count because $a_1$ and $a_2$ belong to the same cluster while $a_4$ belongs to a different one. This leads to the relation $n_{1,4} = n^d_{2,4} + n_{1,4}^d = n_{2,4}$. Consequently, inter-cluster regret can be reduced from a pairwise formulation to a cluster-level representation by avoiding repeated accounting of shared observations. Thus, for $C$ clusters, the cluster-level component of the regret is effectively captured at the cluster level. For the remaining intra-cluster pairs, the regret is determined by the comparisons among the $K_i$ arms within the $i$-th cluster. 
    
    Let $n({c_i, c_j})$ denote the total number of observations of all pairs spanning two clusters $c_i$ and $c_j$. By performing this decomposition, the cumulative regret can be transformed as:
    \begin{equation}
        R(T) \leq \sum_{i,j \in \mathcal{A},~i< j} n^d_{i,j}(T) \leq \sum_{i,j\in\{1, \dots, C\}} n(c_i,c_j) + \sum_{c=1}^{ C}\sum_{i,j\in c_{c},~i < j}n_{i,j}^d.
    \end{equation}
    Note that the inequality sign reflects the potential overlap in the soft-clustering scenario. 
    
    We now study the regret bound for each component. For inter-cluster part, the cumulative regret is bounded as follows:
    \begin{equation}
        \sum_{i,j\in\{1, \dots, C\}, i < j} n(c_i,c_j)  = \frac{C(C-1)}{2} \max_{i,j} D_{i,j}^w \ln  t ~\Rightarrow ~ \mathcal{O}(\frac{1}{\min_{i,j,r} w_{i,j}^r}C^2\ln t).
    \end{equation}
    For the remaining $C$ intra-cluster components, if all regrets are counted, the cumulative regret can be bounded by $\sum_C\frac{K_i(K_i- 1)}{2} \max_{i,j}D^w_{i,j}\ln t$. This bound follows from a microscopic view in which all arms are treated independently withing their clusters. However, in \textit{IPEA-RUCB}, candidates are selected not only based on intra-cluster winning times but also on global winning times. Therefore, for sub-optimal clusters that do not contain the winner $a_*$, the number of times that \texttt{IPEA-RUCB} selects pairs within the same cluster is bounded by a constant smaller than $C(\delta)$, as required to satisfy the concentration condition \cite{ZoghiWMR14}. As a result, the regret arising from intra-cluster comparisons is bounded by $\mathcal{O}(\frac{1}{\min_{i,j,r}w_{i,j}^r}\max_i K_i^2 \ln t)$. Combining both inter- and intra-cluster bounds yields the total cumulative regret bound as $\mathcal{O}(\frac{1}{\min_{i,j,r}w_{i,j}^r}\hat K^2 \ln t)$ where $\hat K = \max\{C, K_1, \dots, K_c\}$.
\begin{figure*}[t]
    \centering
    \includegraphics[width=0.2\linewidth]{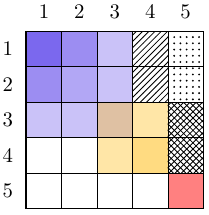}
    \caption{Illustration of soft clustering and regret decomposition. The color-filled regions represent intra-cluster groups: ${a_1, a_2, a_3}$, ${a_3, a_4}$, and the singleton ${a_5}$. The three line-patterned regions in the upper right correspond to inter-cluster comparisons with shared observations across related pairs.}
    \label{fig:regret_illustration}
\end{figure*}
\end{proof}

\section{Broader impacts}
\label{sec:appendix_impact}
Considering the broad development of personalization systems, IPE will play an important role in aligning system behavior with human preferences. With the foundational improvements proposed in our work, personalization systems can more efficiently capture and interpret user requirements, enabling better adaptation to individual needs. This capability is particularly critical in high-stakes applications such as healthcare and transportation, where humans are at the center of the system and alignment with their preferences can have lasting consequences.

At the same time, our work also raises potential negative impacts. The annotation of latent dependencies may introduce fairness and ethical concerns, particularly when such annotations reflect biased or offensive content. Care must be taken to ensure that the integration of contextual signals and richer forms of human feedback does not compromise equity or user trust. 

% \newpage

\section{Limitations}
\label{sec:appendix_limitations}

% We have included a Limitations section in Appendix~\ref{sec:appendix_limitations}, which discusses the limitations of the underlying assumptions, factors affecting performance, and considerations regarding computational efficiency.

The proposed framework offers new insights into the design of DB algorithms, but several limitations remain that present opportunities for future work.

From a theoretical perspective, the current assumption of bidirectional dependency could be further relaxed to support more general dependency structures. Our analysis primarily focuses on deterministic strategies, and a more comprehensive theoretical investigation of stochastic strategies, such as \texttt{IPEA-DTS}, remains an open direction. Additionally, clearer principles for navigating the trade-off between efficiency and augmentation strength, either through refined regret analysis or conditions for uncertainty calibration, would provide stronger algorithmic guidance. For non-stationary preferences, an adaptive design to capture and respond to changes is also required \cite{yu2025learning,yang2025walking}.

From an algorithmic perspective, our prototype design could be enhanced in several ways, including the adoption of task-specific similarity metrics, more flexible in-context inference, and improved annotation strategies. In addition, given the potentially large space of dependencies, developing more efficient strategies and narrowing down the search space for acquiring them could substantially reduce computational costs \cite{WangL24}. Moreover, the availability of benchmarks with richer forms of human feedback remains limited. Building such datasets would provide stronger validation for future research.

Lastly, our framework does not currently include explicit safeguards against annotation bias or misuse of feedback sources. This is particularly relevant when leveraging automated annotations from LLMs. Future work could explore the integration of robustness mechanisms to address these ethical and practical concerns in real-world deployment scenarios.

% \input{appendix_impact}

% \newpage
\section{List of Notations}
For convenience, we list the main notations used in this work in Table~\ref{tab:listofnotations}.
% \onecolumn
\begin{longtable}{>{\raggedright\arraybackslash}p{0.24\linewidth} p{0.70\linewidth}}
\caption{Summary of the Main Mathematical Notation}\label{tab:listofnotations} \\
\hline
\textbf{Symbol} & \textbf{Description} \\
\hline
\endfirsthead
\multicolumn{2}{c}%
{{\bfseries \tablename\ \thetable{} -- continued from previous page}} \\
\hline
\textbf{Symbol} & \textbf{Description} \\
\hline
\endhead
\hline \multicolumn{2}{r}{{Continued on next page}} \\
\endfoot
\hline
\endlastfoot
$K$ & Total number of arms / candidates \\ 
$C$, $c_i$ & Number of cluster with $C \leq K$, and the $i$-th cluster $c_i$\\
$\mathcal{A}$ & Index set of arms from $\left\{1, \dots,  K\right\}$\\
$a_*$ & Winner arm / best candidate \\
$(a_i, a_j)$ & Pair of arm $a_i$ and arm $a_j$, $i,j\in \mathcal{A}$  \\ 
$\mathbf{P} := [p_{i,j}]_{K\times K}$ & Preference matrix with winning probability of each pair $p_{i,j}$ \\ 
$\mathbf{B} :=[b_{i,j}]_{K\times K}$ & Recorded winning matrix with $b_{i,j}$ the number of times $a_i$ beats $a_j$\\
$\tilde p_{i,j} \text{ or } \hat p_{i,j}$ & Estimated mean of $p_{i,j}$, $\hat p_{i,j}$ includes augmented feedback \\
$\mathbf{U} := [\tilde u_{i,j}  \text{ or } \hat u_{i,j}]_{K\times K}$ & Matrix of upper confidence bounds, $\hat u$ denotes augmented bounds\\
$\mathbf{L} := [\tilde l_{i,j} \text{ or } \hat l_{i,j}]_{K\times K}$ & Matrix of lower confidence bounds, $\hat l$ denotes augmented bounds \\
$x_i \in \mathcal{X}$ & Context vector $x_i$ associated with $a_i$ in the context space $\mathcal{X}$ \\
$f(x_i, x_j)$ & Reward function of pair $(a_i, a_j)$ associated with $x_i,~x_j\in \mathcal{X}$ \\
$\theta$  & Unknown parameter in the reward function $f$ \\
$\Delta_i$ & Difference from the best winning probability $p_{*,i}-\frac{1}{2}$ \\
$t$, $T$ & Current iteration $t$ and overall horizon of iterations $T$ \\
$R(T) := \sum_{t=1}^T r_t$ & Cumulative regret over $T$ rounds \\
$\alpha$ & Exploration coefficient \\
$n_{i,j}^d(t) := b_{i,j} + b_{j,i}$ & Number of direct observations of pair $(a_i, a_j)$ up to time $t$ \\ 
$n_{i,j}^r(t)$ & Number of related observations of pair $(a_i, a_j)$ up to time $t$ \\ 
$n_{i,j}(t)$ & Total count of observations (both direct and related) \\
$X_{i,j}^k$ & Random variable of the $k$-th related observation for pair $(a_i, a_j)$ \\
$w_{i,j}^r$ & Dependency from the $i$-th related observation of pair $(a_i, a_j)$ \\
$\eta$ & Ratio $ \eta = n_{i,j}^d + \sum_{i=1}^{n_{i,j}^r} w_{i,j}^k ~/~n_{i,j}(t)$ \\
$\delta$ & Probability in concentration property \\
$C(\delta)$  & Upper bound of time for concentration $C(\delta) =  \left(\frac{(4 \alpha - 1) K^2}{(2 \alpha - 1) \delta}\right)^{\frac{1}{2\alpha - 1}}$ \\
$D_{i,j}^w$ & $D_{i,j}^w = {4\alpha} \cdot ({\min_r w_{i,j}^{r~2} \cdot \min\{\Delta_i^2, \Delta_j^2\} })^{-1}$ \\
$K_i$ & The $i$-th subset, $i \in \{1, \dots, C\}$ \\
$\hat K$ & $\hat K = \max\{ C, K_1, \dots, K_C \}$ \\
$W$ & Dictionary of related observations \\
$\mathcal{G}$ & Similarity graph from context
\end{longtable}
% \twocolumn

\end{document}